\documentclass{article}

\PassOptionsToPackage{numbers}{natbib}

    \usepackage[preprint]{neurips_2025}

\usepackage[utf8]{inputenc} 
\usepackage[T1]{fontenc}    
\usepackage{hyperref}       
\usepackage{url}            
\usepackage{booktabs}       
\usepackage{amsfonts}       
\usepackage{nicefrac}       
\usepackage{microtype}      
\usepackage{xcolor}         

\usepackage{graphicx,subcaption}
\usepackage{enumitem}
\usepackage{algorithm}
\usepackage{algorithmic}
\usepackage{booktabs}  
\usepackage{ragged2e}
\usepackage{wrapfig}
\usepackage{arydshln}
\usepackage{setspace}
\usepackage{adjustbox} 
\usepackage{framed}

\usepackage{amssymb}
\usepackage{mathtools}
\usepackage{amsthm}

\newbool{makeBlue}
\setbool{makeBlue}{false} 
\newcommand{\conditionalblue}[1]{%
  \ifbool{makeBlue}{%
    \textcolor{blue}{#1}%
  }{%
    #1%
  }%
}

\theoremstyle{plain}
\newtheorem{theorem}{Theorem}[section]

\theoremstyle{definition}
\newtheorem{definition}[theorem]{Definition}

\theoremstyle{remark}

\title{Balancing Multimodal Training Through Game-Theoretic Regularization}

\author{%
  Konstantinos Kontras$^{1}$\thanks{Correspondence to: \texttt{konstantinos.kontras@kuleuven.be}} \\ \And
  Thomas Strypsteen$^{1}$ \\ \And
  Christos Chatzichristos$^{1}$ \\ \And
  Paul Pu Liang$^{3}$ \\ \And
  Matthew Blaschko$^{1}$ \\ \And
  Maarten De Vos$^{1,2}$ \\\And
  \\
  $^{1}$Department of Electrical Engineering, KU Leuven, Leuven, Belgium \\
  $^{2}$Department of Development and Regeneration, KU Leuven, Leuven, Belgium \\
  $^{3}$Media Lab and EECS, MIT, Boston, USA 
}

\begin{document}

\maketitle

\begin{abstract}


Multimodal learning holds promise for richer information extraction by capturing dependencies across data sources. Yet, current training methods often underperform due to modality competition, a phenomenon where modalities contend for training resources leaving some underoptimized. This raises a pivotal question: how can we address training imbalances, ensure adequate optimization across all modalities, and achieve consistent performance improvements as we transition from unimodal to multimodal data? This paper proposes the Multimodal Competition Regularizer (MCR), inspired by a mutual information (MI) decomposition designed to prevent the adverse effects of competition in multimodal training. Our key contributions are: 
1) A game-theoretic framework that adaptively balances modality contributions by encouraging each to maximize its informative role in the final prediction
2) Refining lower and upper bounds for each MI term to enhance the extraction of both task-relevant unique and shared information across modalities.
3) Proposing latent space permutations for conditional MI estimation, significantly improving computational efficiency.
MCR outperforms all previously suggested training strategies and simple baseline, clearly demonstrating that training modalities jointly leads to important performance gains on both synthetic and large real-world datasets. We release our code and models at {\color{blue}https://github.com/kkontras/MCR}.
\end{abstract}

\section{Introduction}

Exploiting multimodal data has made significant progress, with advances in generalizable representations and larger datasets enabling solutions to previously unattainable tasks \cite{kontras2024core, liang2024foundations, li2022blip, nagrani2021attention, radevski2023multimodal, radevski2021revisiting, radford2021learning, tsai2019multimodal, wang2017adversarial, zadeh2017tensor}. However, studies indicate that jointly trained multimodal data is often utilized suboptimally, underperforming compared to ensembles of unimodal models, jointly trained modalities, or even the best single modality \cite{wang2020makes, MSLR}. The expectation that adding a new modality should improve performance, assuming independent errors and above-chance predictive power \cite{hansen1990neural}, is frequently contradicted in practice.

\begin{figure*}[t]
    \centering
    \includegraphics[width=0.99\linewidth]{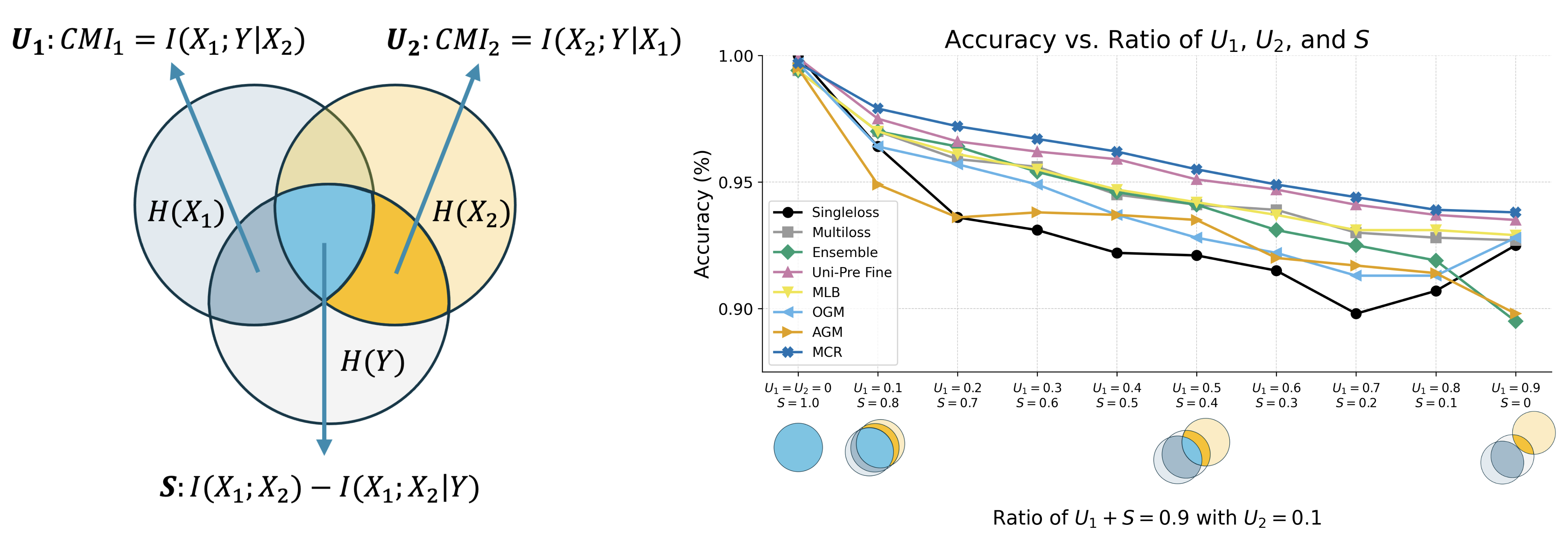}
    \caption{
    (Left) Illustration of the conditional mutual information ($\operatorname{CMI}$) terms, $\operatorname{CMI}_1: I(X_1; Y \mid X_2)$ and $\operatorname{CMI}_2: I(X_2; Y \mid X_1)$, representing the unique contributions ($U_1$, $U_2$) of each modality. The shared task-relevant information ($S$) is defined as $I(X_1; X_2) - I(X_1; X_2 \mid Y)$. (Right) 
    Accuracy on a synthetic dataset designed to induce \textbf{multimodal competition}. We vary the ratio of unique information from modality 1 ($U_1$) to shared information ($S$), while keeping the contribution of modality 2 ($U_2$) constant. As the imbalance increases (moving right on the x-axis), the performance of most methods drops. The standard Joint Training (Singleloss) approach shows a steep decline, highlighting its vulnerability to modality competition where one modality dominates and suppresses the other. In contrast, our method, $\operatorname{MCR}$, demonstrates greater robustness by maintaining the highest accuracy and exhibiting the slowest performance degradation. See Section \ref{sec:synthetic_results} for more details. 
    }
    \label{fig:synthetic_intro}
\end{figure*}

\citet{huang2022modality} attribute this issue to modality competition, where one modality quickly minimizes training error, misdirecting and suppressing the learning of others. To counteract this effect, monitoring each modality's contribution during training and applying corrective measures is crucial. To this end, several balancing strategies have been proposed \cite{du2023uni, PMR, fujimori2020modality, gat2020removing, ji2022increasing, MLB, AGM, OGMGE, xsleepnet, wang2020makes, wu2022characterizing, MSLR, MMPareto, wei2024diagnosing, zhang2024multimodal, ReconBoost, wei2024enhancing}. Some ignore a modality's contribution beyond its independent unimodal performance, while others address this by measuring output differences under input perturbation, but at the cost of increased sensitivity to these perturbations and significant computational overhead. Moreover, it is crucial to examine whether enhancing one modality's influence on the output does not come at the expense of others, as this could undermine overall performance.

\textit{
\indent Given these challenges, how can we efficiently regularize multimodal competition to ensure balanced and effective learning across modalities?}

This paper introduces a loss function encouraging the exploration of task-relevant information across modalities, the \textsc{Multimodal Competition Regularizer ($\operatorname{MCR}$)}. The approach incorporates the following key contributions:
\vspace{-0.2cm}
\begin{enumerate}[leftmargin=0.3cm, labelsep=0.1cm, itemsep=0.0cm]
    \item \textbf{MI Bounds:} We decompose joint mutual information into task-relevant shared and unique components, using refined lower and upper bounds to promote informative signals and suppress noise.
    
    \item \textbf{Game-Theoretic Modality Balancing:} We frame modality interaction through a game-theoretic framework, allowing each modality to adjust its contribution throughout training.
    
    \item \textbf{Efficient CMI Estimation:} We introduce latent-space perturbations for low-cost conditional MI estimation, avoiding repeated full-model passes.
\end{enumerate}

\vspace{-0.2cm}
We extensively evaluate $\operatorname{MCR}$ on synthetic datasets and several established real-world multimodal benchmarks, including action recognition on AVE \cite{tian2018audio} and UCF \cite{soomro2012ucf101}, emotion recognition on CREMA-D \cite{cao2014crema}, human sentiment on CMU-MOSI \cite{zadeh2016mosi}, human emotions on CMU-MOSEI \cite{zadeh2018multimodal}, and egocentric action recognition on Something-Something \cite{goyal2017something}. Our results demonstrate that $\operatorname{MCR}$ outperforms all previous methods and simple baselines across various datasets and models, improving multimodal supervised training.

\section{Problem Analysis and Related Work}
\label{sec:problem_formulation}


Consider a dataset of \(N\) independent and identically distributed (i.i.d.) datapoints sampled from a distribution \(\mathcal{D}\), where each datapoint has \(M\) modalities \(X = (X_1, \dots, X_M)\) and a target \(Y_t\). Our goal is to learn a parameterized function \(f: X; \theta \rightarrow Y_t\), where \(\theta\) denotes all the model's learnable parameters. The unimodal encoder for each modality is defined as \(f_m: X_m; \theta_m \rightarrow Z_m\), encoding input \(X_m\) into a latent representation \(Z_m\). The fusion network \(f_c: [Z_1, \dots, Z_M]; \theta_c \rightarrow Y_t\) predicts \(Y_t\) from the latent representations, as do the unimodal task heads \(f_{c_m}: Z_m; \theta_{c_m} \rightarrow Y_m\). Model families are defined as, unimodal models for $m=[1, .. ,M]$ modalities:
\begin{equation}
\label{eq:unimodal_model}
    \mathcal{F}_{u_m}: 
f_{u_m}\left(X_{m}; \theta_{m}, \theta_{c_m} \right) = f_{c_m}\left(f_{m}\left(X_{m}; \theta_{m}\right); \theta_{c_m}\right),
\end{equation}
and for multimodal models:
\begin{equation}
    \mathcal{F}: f(X; \theta) = f_c\left(\left[f_{1}\left(X_{1}; \theta_{1}\right), .., f_{M}\left(X_{M}; \theta_{M}\right)\right]; \theta_c\right).
\end{equation}
For simplicity, we continue our analysis with \(M=2\), focusing on models with two modalities.

\subsection{The limitation of supervised multimodal training}

In supervised learning, the goal is to learn representations \( Z_1 =  f_1(X_1;\theta_1) \) and \( Z_2 =  f_2(X_2;\theta_2) \) when fused via \( f_c(\left[Z_1, Z_2\right]) \), yield accurate predictions. This is achieved by minimizing the task loss or, equivalently, by maximizing the MI between the fused representation and the target:
\begin{equation}
\label{eq:supervised_mutual}
\mathop{\arg\max}_{\substack{Z_1 := f_1(X_1;\theta_1),\\ Z_2 := f_2(X_2;\theta_2)}} I(f_c(\left[Z_1, Z_2\right]); Y_t).
\end{equation}

During training, models often over-rely on the stronger or more accessible modality, limiting the contribution of others. This leads to mutual information being dominated by one modality, e.g., \(I(f_c(\left[Z_1, Z_2\right]); Y) \approx I(Z_1; Y)\) with \( I(Z_2; Y \mid Z_1) \approx 0 \), indicating that $Z_2$ adds little once $Z_1$ is learned. See Appendix \ref{sec:evidence_of_supervised_limitations} for an illustrative experiment and Appendix \ref{app:MCE} for a formal definition of the resulting generalization gap.



This kind of imbalance is well-known in single-modality learning, where dominant features can overshadow others, harming generalization. Regularization techniques like $l_1 \,/ \, l_2$ penalties and dropout promote balanced feature use \cite{ng2004feature, srivastava2014dropout}, but their adaptation to multimodal settings is nontrivial. For example, applying modality-specific dropout \cite{xiao2020audiovisual} offers limited benefits \cite{OGMGE}. The core challenge remains: how to effectively regulate interaction and competition between modalities.

\subsection{Related Work}

Prior research has explored various strategies for multimodal learning, ranging from simple unimodal and ensemble-based approaches to more sophisticated methods for balancing modality contributions. Unimodal training optimizes each modality separately, while ensemble methods combine unimodal predictions without additional training. Joint training optimizes all modalities under a single-loss objective but does not explicitly ensure sufficient training for each modality. To address this, Multi-Loss \cite{vielzeuf2018centralnet} introduces additional unimodal task losses, and MMCosine \cite{xu2023mmcosine} equalizes modality influence by standardizing features and weights. Pre-trained unimodal encoders are often used, either with frozen weights (Uni-Pre Frozen) or fine-tuned jointly (Uni-Pre Finetuned). Other adaptive strategies include MSLR \cite{MSLR}, which adjusts learning rates based on unimodal validation performance, OGM \cite{OGMGE}, which modulates gradients by comparing unimodal performance across modalities, and MLB \cite{MLB}, which combines unimodal task losses and modulates gradients from both unimodal and multimodal objectives, MMPareto \cite{MMPareto} that mitigates gradient conflicts between modalities by equalizing the contribution of unimodal and multimodal gradient and D\&R \cite{wei2024diagnosing} suggest a new strategy where modalities that overfit get their part of the network partially reweighted with the initial weights of the training. 

Most of these methods assume distributional independence and measure modality contributions through unimodal performance, which can be a limited indicator, missing cases where modality correlation is crucial. Other approaches estimate influence based on prediction differences after perturbations \cite{AGM, ji2022increasing, gat2020removing}. AGM \cite{AGM} uses zero-masking Shapley values directly optimizing them as unimodal predictors,  \citet{wei2024enhancing} use a permutation-based Shapley values and resampling of the training set to affect the training, while other methods address similar problems by introducing perturbations such as Gaussian noise \cite{gat2020removing} or task-specific augmentations \cite{ji2022increasing, liang2024factorized}. However, perturbation-based approaches increase the network's sensitivity to the chosen perturbations and hinder scalability due to their higher computational demands.

A line of work keeps unimodal training as the primary strategy. MLA \cite{zhang2024multimodal} uses a shared task head and dynamic weighted summation during validation, while ReconBoost \cite{ReconBoost} alternates unimodal updates with agreement and diversity regularization before finetuning the ensemble. However, these approaches avoid multimodal training in the earlier steps to mitigate conflicts, yet overlook the potential benefits of direct multimodal interactions in those steps. 
\section{Multimodal Competition Regularizer}

\begin{figure*}
    \centering
    \includegraphics[width=0.9\linewidth]{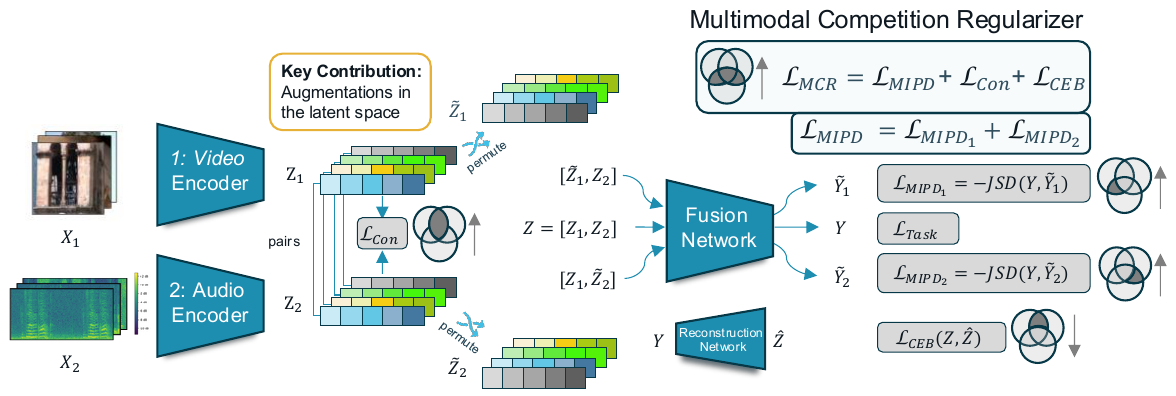}
    \caption{
\textbf{Multimodal Competition Regularizer ($\operatorname{MCR}$):} The diagram illustrates the $\operatorname{MCR}$ framework, which mitigates modality competition in multimodal learning. Raw data ($X_1$ and $X_2$) are encoded into latent representations ($Z_1$ and $Z_2$), which are then permuted to create $\tilde{Z}_1$ and $\tilde{Z}_2$ and the paired combinations. These combinations are passed through the Fusion Network to produce predicted outputs ($Y$, $\tilde{Y}_1$, $\tilde{Y}_2$). The comparison between predictions reveals each modality's contribution. For example, if $Y \approx \tilde{Y}_1$, it shows that $X_1$ has little impact, and the model relies on $X_2$. The $\operatorname{MCR}$ loss includes three components: $\mathcal{L}_{\operatorname{MIPD}}$ maximize the Jensen-Shannon divergence (JSD) between task output and permuted modality predictions. $\mathcal{L}_{\operatorname{Con}}$ aligns modality representations, while $\mathcal{L}_{\operatorname{CEB}}$ penalizes task-irrelevant information by reconstructing back to the latent space. 
}
    \label{fig:concept}
\end{figure*}

Multimodal competition arises when a model trained on multiple modalities prioritizes one, leading to over-reliance and reducing the contribution of others. This imbalance limits the model’s ability to fully utilize all available information. In this section, we introduce $\mathcal{L}_{\operatorname{MCR}}$, a set of loss components designed to address multimodal competition. Each component of the loss is motivated by the following MI decomposition:
\begin{align}
I(X_{1}; X_{2}; Y) =&
\underbrace{I( X_1; Y \mid X_2) + I( X_2; Y \mid X_1)}_{ 
\substack{ 
\text{Task-Relevant Unique Information} \\ 
\text{of each modality} \sim \mathcal{L}_{ \operatorname{MIPD}}} }
+ \hspace{-3pt} \underbrace{I(X_1; X_2)}_{
\substack{ 
\text{Shared Information} \\ 
\mathcal{L}_{\operatorname{Con}}}} \hspace{-3pt}
- \hspace{-8pt} \underbrace{I(X_1; X_2 \mid Y)}_{
\substack{ 
\text{Task-Irrelevant} \\ 
 \text{Shared Information} \sim  \mathcal{L}_{\operatorname{CEB}}}
 }\label{eq:mutual_decomposition}\hspace{-10pt}.
\end{align}
This decomposition is illustrated by the Venn diagram in Figure \ref{fig:synthetic_intro}. The \(\operatorname{CMIs}\) $I( X_1; Y \mid X_2)$ and $I( X_2; Y \mid X_1)$ capture modality-specific information for predicting the target. Maximizing them with the Mutual Information Perturbed Difference ($\operatorname{MIPD}$) loss, $\mathcal{L}_{\operatorname{MIPD}}$, which assesses each modality's contribution via output variations under input perturbations (elaborated in Sec. \ref{sec:permutations}) and encourages the extraction of modality-specific, task-relevant features. The third term, $I(X_1; X_2)$, quantifies shared information between modalities. Maximizing it with a contrastive loss, $\mathcal{L}_{\operatorname{Con}}$, aligns representations and leverages their shared information effectively \cite{jia2021scaling, oord2018representation, radford2021learning}. The final term, $I(X_1; X_2\mid Y)$, represents task-irrelevant shared information. Penalizing it with the conditional entropy bottleneck ($\operatorname{CEB}$) \cite{fischer2020ceb} and the corresponding loss $\mathcal{L}_{\operatorname{CEB}}$ to filter out irrelevant information, focusing the model on features relevant to the downstream task. Each term has a corresponding loss, as illustrated in Figure \ref{fig:concept}, forming the regularizer with three key losses:
\begin{equation}
\label{eq:total}
 \scalebox{1.2}{$\mathcal{L}_{\operatorname{MCR}} = \mathcal{L}_{\operatorname{MIPD}} + \mathcal{L}_{\operatorname{Con}} + \mathcal{L}_{\operatorname{CEB}}$}
\end{equation}


\begin{figure*}[t]
    \centering
    \includegraphics[width=0.7\linewidth]{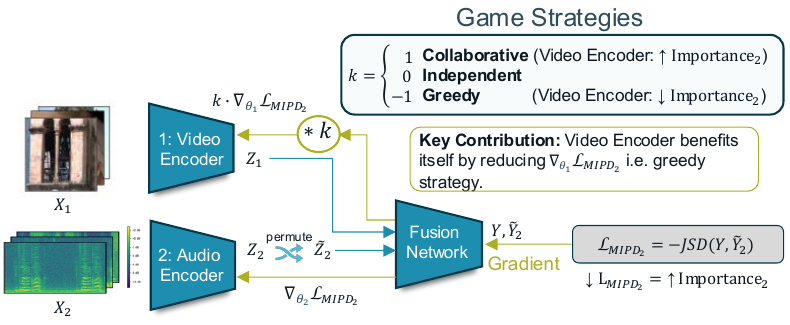}
    \caption{
This figure illustrates a key aspect of our training process, showing how competition strategies between modalities are applied. The gradient multiplier adjusts the video encoder's response to Audio Importance ($\text{Importance}_2$). When $k=1$, the video encoder enhances $\text{Importance}_2$; at $k=0$, it remains neutral, and at $k=-1$, it competes by reducing $\text{Importance}_2$ to prioritize its own ($\text{Importance}_1$). This reflects the principle that increasing the importance of one modality can reduce the importance of the other.}
    \label{fig:game_strategies}
\end{figure*}

\subsection{Approximating MI Terms}

$\mathbf{I( X_1; Y \mid X_2):}$ To approximate each \(\operatorname{CMI}\) and capture the unique contribution of each modality, the \(\operatorname{MIPD}\) serves as a surrogate function, measuring how input perturbations affect the model’s output. By comparing predictions with and without these perturbations, \(\operatorname{MIPD}\) estimates how much information each modality provides. If a modality is crucial, altering its input should significantly change the output, revealing its importance. 

Estimating the \(\operatorname{CMI}\) directly through $I(X_1; Y \mid X_2)= H(Y\mid X_2) - H(Y\mid X_1, X_2)$ is typically intractable. Instead we use the $\operatorname{MIPD}$ as a lower bound, defined as:
\begin{align} 
\operatorname{MIPD}(X_{1}; Y \mid X_{2}) &= I(X_{1}; Y \mid X_{2}) - I(\tilde{X}_{1}; Y \mid X_{2}) \leq I(X_{1}; Y \mid X_{2}), 
\end{align}
where the perturbed version of modality $X_1$ is denoted as $\tilde{X}_{1}$. Interpreting MI via entropy, each $\operatorname{CMI}$ can be expressed as the difference of the $\log$ probabilities with and without the perturbations:  
\begin{align}
    \operatorname{MIPD}(X_{1}; Y \mid X_{2})
   &= H(Y \mid X_{2}, \tilde{X}_{1}) - H(Y \mid X_{2}, X_{1}) \notag 
   \\ 
    &= \hspace{-28pt}
  \underset{\raisebox{-7pt}{$\hspace{10pt} \scriptstyle \substack{y \sim p(y)\\ \hspace{7pt}
    x_1, x_2 \sim p(x_1, x_2, y)}$}}{\mathbb{E}}
    \hspace{-23pt}
\left[    - \hspace{-14pt}
   \underset{\raisebox{-2pt}{$\hspace{10pt}\scriptstyle \substack{\tilde{x}_1 \sim p(x_1)}$}}{\mathbb{E}}
\hspace{-12pt}\left[ \log p(y \mid x_{2}, \tilde{x}_{1}) \right] + \log p(y \mid x_{2}, x_{1}) \right]. 
\raisetag{8pt}
\end{align}
Instead of this log-likelihood ratio, we use the symmetrically bounded Jensen-Shannon divergence (\(\operatorname{JSD}\)) \cite{lin1991divergence} to prevent training instabilities, leading to the following:
\begin{align}
    \begin{split}
    \mathcal{L}_{\operatorname{MIPD_{1}}} &=  -\operatorname{MIPD}(X_{1}; Y \mid X_{2})  
    = - \hspace{-30pt}
   \underset{\raisebox{-10pt}{$\hspace{10pt}\scriptstyle \substack{y \sim p(y) \\ \hspace{7pt}x_1, x_2 \sim p(x_1, x_2, y) \\ \tilde{x}_1 \sim p(x_1)}$}}{\mathbb{E}}
\hspace{-25pt}
\left[ \operatorname{JSD}( p(y \mid x_{2}, x_{1}), p(y \mid x_{2}, \tilde{x}_{1})) \right].
   \label{eq:LMIPD}
\end{split}
\raisetag{17pt} 
\end{align}
Similarly, $\mathcal{L}_{\operatorname{MIPD}_{2}}$ can be computed symmetrically. 

$\mathbf{I( X_1; X_2):}$ The next \(\operatorname{MI}\) term measures how much information the two modalities share, capturing the common patterns between the modalities and aligning the representations of these shared aspects. We exploit the available label information employing the supervised contrastive loss $\mathcal{L}_{\operatorname{Con}}$ \cite{khosla2020supervised}:
\begin{align}
\label{eq:contrastive}
\mathcal{L}_{\operatorname{Con}} &= \hspace{-15pt}
   \underset{\raisebox{-10pt}{$\hspace{10pt}\scriptstyle \substack{
       x_1, y \sim p(x_1, y) \\
    \hspace{5pt} x_2^+ \sim p(x_2\mid y) \\
    \hspace{10.5pt} x_2^- \sim p(x_2\mid \neg y)\footnotemark
   }$}}{\mathbb{E}}
\hspace{-4pt}\left[ \log \frac{ \psi(x_1, x_2^+)}{\sum_k  \psi(x_1, x_{2_k}^-)} \right],
\raisetag{15pt}
\end{align}
\footnotetext{\(\neg y = \{y' \in \mathcal{Y} \mid y' \neq y\}\), where \(\mathcal{Y}\) is the set of target labels.}
where $\psi$ is the critic function, which, in our case, is the exponential dot product. Minimizing the $\mathcal{L}_{\operatorname{Con}}$, maximizes a lower bound on both the  \(\operatorname{MI}\) between the two modalities and the $\operatorname{CMI}$ terms:
\begin{align}
I(X_{2}; Y|X_{1}) + I(X_{1}; Y|X_{2}) + 2I(X_{2}; X_{1}) \geq \log N - \mathcal{L}_{\operatorname{Con}}^{Opt}.
\end{align}
As $N$ increases, the bound becomes tighter, while the bound is not affected by the number of positive samples (same class datapoints). More details are provided in Appendix \ref{sec:app_Con_proof}. 

$\mathbf{I(X_{1}; X_2 \mid Y):}$ The final term captures irrelevant shared information between modalities, and minimizing an upper bound on this ensures the model retains only task-relevant content. For this purpose, we exploit the idea of Conditional Entropy Bottleneck ($\operatorname{CEB}$) $L_{\operatorname{CEB}}$ \cite{fischer2020conditional}, targeting superfluous information in multimodal representations via a reconstruction loss. A small reconstruction head, \( h: Y; \theta_h \rightarrow Z=(Z_1,Z_2) \), predicts back the latent space, effectively filtering out irrelevant content:
\begin{align}
    \mathcal{L}_{\operatorname{CEB}} = \hspace{-35pt}
       \underset{\raisebox{-5pt}{$\hspace{10pt}\scriptstyle \substack{ x_1, x_2, y \sim p(x_1, x_2, y)}$}}{\mathbb{E}}
\hspace{-35pt}\left\| [f_1(x_1),f_2(x_2)] - h(y ; \theta_h ) \right\|^2
\raisetag{0.2cm}
\label{eq:CEB}\end{align}
The exact derivation of this loss term can be found in Appendix~\ref{sec:app_ceb}. Penalizing irrelevant information has been shown to enhance calibration and robustness \cite{fischer2020ceb}, but it must be carefully evaluated, as it can introduce constraints that may hinder overall performance.

\subsection{The Game of Multimodal Fusion}



We adopt a game-theoretic approach to balance the terms of the proposed $\mathcal{L}_{\operatorname{MIPD}}$. The key idea is that increasing one modality's importance (e.g., via $\operatorname{MIPD}_1$) can inherently reduce the other's (e.g., $\operatorname{MIPD}_2$). Thus, an underutilized encoder $i$ (with parameters $\theta_i$) can boost its relevance both by minimizing $\mathcal{L}_{\operatorname{MIPD}_i}$ and by maximizing $\mathcal{L}_{\operatorname{MIPD}_{\neg i}}$. This twofold strategy helps prevent suppression of weaker modalities. We frame $\mathcal{L}_{\operatorname{MIPD}}$ as a game where each encoder (player) selects a strategy, minimize, maximize, or ignore. Figure \ref{fig:game_strategies} illustrates how the video modality, via a hyperparameter $k$, can choose to assist, ignore, or diminish the audio modality. Each encoder applies this logic selectively as formalized below:
\begin{align}
\nabla_{\theta_1} \mathcal{L}_{\operatorname{MIPD}}
&= \lambda_M \left( 
\nabla_{\theta_1}\mathcal{L}_{\operatorname{MIPD}_1} 
+ k \, \nabla_{\theta_1}\mathcal{L}_{\operatorname{MIPD}_2}
\right), \\
\nabla_{\theta_2} \mathcal{L}_{\operatorname{MIPD}}
&= \lambda_M \left( 
\nabla_{\theta_2}\mathcal{L}_{\operatorname{MIPD}_2} 
+ k \, \nabla_{\theta_2}\mathcal{L}_{\operatorname{MIPD}_1}
\right). \footnotemark
\label{eq:competition_terms}
\end{align}
\footnotetext{The parameter set under the loss indicates where backpropagation applies.}
where $\lambda_M$ is a Lagrange multiplier, and $k \in \{-1, 0, 1\}$ sets the modality’s strategy:
\begin{itemize} 
[leftmargin=0.3cm, labelsep=0.1cm, itemsep=0.0cm]

\item \textbf{Collaborative ($k=1$):} All modalities work together to increase each other's contributions. The $\mathcal{L}_{\operatorname{MIPD}}$ terms are applied across all parameters, resulting in $\min\limits_{\theta}\mathcal{L}_{\operatorname{MIPD}}$.

\item \textbf{Independent ($k=0$):} Each modality focuses on maximizing its own contribution by optimizing solely its respective $\mathcal{L}_{\operatorname{MIPD}}$ term, leading to $\min\limits_{\theta_i}\mathcal{L}_{\operatorname{MIPD}_{i}}$.

\item \textbf{Greedy ($k=-1$):} Each modality seeks to maximize its own contribution by: 1) minimizing its own $\mathcal{L}_{\operatorname{MIPD}}$ term, and 2) maximizing the $\mathcal{L}_{\operatorname{MIPD}}$ terms of other modalities, resulting in a min-max game, $\min\limits_{\theta_i} \max\limits_{\theta_{\neg i}} \mathcal{L}_{\operatorname{MIPD}_{i}}$\footnote{The notation $\neg i$ refers to the rest of the modalities except $i$.}.
\end{itemize}
Following the results in Appendix \ref{app:ablation_study_strategy}, we adopt the greedy strategy as default, as it showed the most consistent performance in our setting.
\vspace{-1mm}

\subsection{Perturbations}
\label{sec:permutations} 
\vspace{-1mm}

To assess the importance of modality $X_1$, we define $\mathcal{L}_{\operatorname{MIPD_{1}}}$, which captures changes in the model's output when $X_1$ is perturbed (i.e., $\{\tilde{X}_1, X_2\}$ vs. $\{X_1, X_2\}$). Instead of traditional input-space perturbations, which can be computationally expensive and task-dependent, we apply a within-batch permutation \(\sigma_{e} \sim \text{Uniform}(\mathcal{P})\) in the latent space, yielding $\tilde{X}_1 = \sigma_{e}(X_1)$. This approach avoids extra forward passes and reduces computational and memory overhead. Further analysis of this technique and comparisons with prior methods are provided in Appendix \ref{app:perturbations} and \ref{app:computational_analysis}.

The complete algorithm is presented in Algorithm~\ref{alg:1}, with an extension of $\mathcal{L}_{\operatorname{MCR}}$ to $M$ modalities described in Appendix~\ref{sec:MIPD_N_Mod}. In Appendix \ref{app:ablation_study_loss}, we analyze various combinations of loss components, revealing that penalizing task-irrelevant information benefits models with extensive SSL pretraining but proves detrimental for those without it.

\begin{algorithm}[h]
\setstretch{1.1}
\caption{Multimodal Training with MCR}
\textbf{Input:} Training dataset \(D\) with modalities \(X_1, X_2, \ldots, X_M\), labels \(Y_t\), multimodal model \(f \in \mathcal{F}\), initialized unimodal encoders \(\theta_i\), reconstruction model \(h\), \(\lambda_{\text{uni}}, \lambda_{\text{M}}\) Lagrangian coefficients:

\begin{algorithmic}[1]
\FOR{each batch \((X_1, .., X_M, Y_t)\) of each epoch}
        \STATE Compute $\mathcal{L}_{task}(f(X_1,.., X_M), Y_t)$ and  $\mathcal{L}_{task}^{uni}= \lambda_{\text{uni}} \sum_{m=1}^{M} \mathcal{L}_{task}(f_m^u(X_m), Y_t)$
        \STATE Extract \( (Z_1, .., Z_M)\) from \(f(X_1, .., X_M)\)
        \STATE Assess the \(\mathcal{L}_{\text{Con}}\) with Eq. \ref{eq:contrastive} and \(\mathcal{L}_{\text{CEB}}\) with Eq. \ref{eq:CEB}
        \STATE Sample \(\sigma_e\) permutations and compute permuted pairs on the latent space $Z$
        \STATE Pass each pair through the fusion model \(f_c\) to get predictions \(\tilde{Y}_{m}\) with modality $m$ permuted
        \STATE Compute \(\mathcal{L}_{\text{MIPD}} \) using Eq. \ref{eq:competition_terms}, \ref{eq:LMIPD} and selecting $k$ by strategy (default: Greedy)
        \STATE Determine $\mathcal{L}_{\operatorname{MCR}}$ from Eq. \ref{eq:total}
        \STATE Update the model parameters based on \\ $ \mathcal{L}_{total} = \mathcal{L}_{task} + \mathcal{L}_{uni} + \mathcal{L}_{\operatorname{MCR}}$ 
\ENDFOR
\end{algorithmic}
\label{alg:1}
\end{algorithm}

\section{Experiments}


\subsection{Synthetic Dataset}
\label{sec:synthetic_results}

We create a scenario where mutual information varies, showcasing modality competition. While various factors can contribute to such a phenomenon, we focus on modality informativeness imbalance to motivate our approach.

\textbf{Data:} We generate task-irrelevant information for each modality by sampling \( N_{1}, N_{2} \sim \mathcal{N}(0, \mathbf{I}) \) and the 5-class label $Y_t$ from a uniform distribution \( Y_t \sim \text{Uniform}(5) \). Each modality is converted into a high-dimensional vector using fixed transformations, similar to Liang et al.\cite{liang2024factorized}. We relate both modalities to the label through a linear relationship: $X_1 = N_{1} + Y_t$ and $X_2 = N_{2} + Y_t$. Data points are distributed in such a way that either both modalities contain label information (Shared Information) or only one of the modalities (Unique Information). In cases where only one modality contains label information, the other modality is defined as $X_1 = N_{1}$ and $X_2 = N_{2}$ respectively. We vary the percentage of data points with shared and unique information to analyze model performance under different conditions.

\textbf{Results}: Figure \ref{fig:synthetic_intro} shows the performance on synthetic data, comparing our method ($\operatorname{MCR}$) with several baselines. As the shared information $S$ among the modalities decreases, and the unique information of one modality $U_1$ increases while the $U_2$ remains constant, we observe a performance drop for all methods. $\operatorname{MCR}$ maintains the highest accuracy across all combinations, demonstrating the slowest decline and highlighting its robustness to such imbalance.

\subsection{Real-World Datasets}

\textbf{Datasets:} We explore several real-world datasets, primarily with video, optical flow, audio, and text modalities, that either exhibit significant imbalance among modalities or serve as standard multimodal benchmarks. Detailed descriptions are provided in Appendix \ref{app:dataset}, with brief summaries below:
\begin{description}[leftmargin=0.0cm, labelsep=0.0cm]
\vspace{-6pt}
    \setlength\itemsep{-0.0em} 
    \raggedright
    \item 1. CREMA-D \cite{cao2014crema}: An emotion recognition dataset with 91 actors expressing 6 distinct emotions.
    \item 2. AVE \cite{tian2018audio}: A collection of videos with temporally aligned audio-visual events across 28 categories.
    \item 3. UCF \cite{soomro2012ucf101}: An action recognition dataset of real-life YouTube videos.
    \item 4. CMU-MOSEI \cite{zadeh2018multimodal}: Multimodal sentiment analysis dataset with 23k monologue clips.
    \item 5. CMU-MOSI \cite{zadeh2016mosi}: Multimodal sentiment analysis dataset with over 2k YouTube video clips.
    \item 6. Something-Something (V2) \cite{goyal2017something}: 220k clips of individuals performing 174 hand actions.
\end{description}

\textbf{Models:} We employ a variety of models and backbone encoders to examine the behavior of both smaller-scale models trained from scratch and larger, more complex models pretrained with self-supervised learning (SSL). This combination demonstrates that our method is effective in both limited data scenarios without pretraining and in cases with ample data where the goal is fine-tuning. We utilize ResNet-18 and small-scale Transformers (from thousand to 20M parameters) alongside state-of-the-art models such as Swin-TF \cite{liu2021swin} and Conformer \cite{goncalves2023versatile}, incorporating backbone encoders like Wav2Vec2, HuBERT, and ViViT, resulting in model sizes approaching 200M parameters. Detailed model configurations for each dataset are provided in Appendix \ref{sec:unimodal_encoders} and experimental details in Appendix \ref{sec:experimental_details}. These choices aim to bridge the gap between theoretical work and practical application.

\begin{table}[h]
\captionsetup{position=above} 
\caption{
Performance comparison of $\operatorname{MCR}$ against prior multimodal training methods across six datasets. $\operatorname{MCR}$ consistently achieves top results across various modality combinations (V-A, V-T, V-A-T, V-OF) and model architectures. MOSI and MOSEI use three modalities and are trained as regression tasks (converted to binary accuracy); the rest are classification tasks. All baselines were rerun under our evaluation protocol to ensure fair comparison and address data leakage issues identified in previous evaluation setups.
}
\label{table:perf_results_mosei_sth}
\centering
\Large
\resizebox{\textwidth}{!}{
\begin{tabular}{lcccccccccc}
\toprule[1.5pt]
& \multicolumn{2}{c}{CREMA-D} & \multicolumn{2}{c}{AVE} & \multicolumn{1}{c}{UCF} & \multicolumn{2}{c}{MOSI} & \multicolumn{2}{c}{MOSEI}&  \multicolumn{1}{c}{Sth-Sth}\\
& \multicolumn{1}{c}{ResNet} & \multicolumn{1}{c}{Conformer}& \multicolumn{1}{c}{ResNet} & \multicolumn{1}{c}{Conformer} & \multicolumn{1}{c}{ResNet}& \multicolumn{4}{c}{Transformer} & \multicolumn{1}{c}{Swin-TF}  \\
Method          & V-A & V-A & V-A & V-A & V-A & V-T & V-A-T & V-T & V-A-T & V-OF \\ \midrule[1.2pt]
Unimodals  & \multicolumn{1}{c}{\begin{tabular}{c} V:  55.4{\tiny$\pm$3.0}\\ A: 60.6{\tiny$\pm$2.3} \end{tabular}}& 
\multicolumn{1}{c}{\begin{tabular}{c} V: 69.4{\tiny$\pm$2.0} \\ A: 76.0{\tiny$\pm$2.6} \end{tabular}}& 
\multicolumn{1}{c}{\begin{tabular}{c} V: 45.7{\tiny$\pm$1.6} \\ A: 62.6{\tiny$\pm$0.9} \end{tabular}}& 
\multicolumn{1}{c}{\begin{tabular}{c} V: 75.5{\tiny$\pm$1.2} \\ A: 76.5{\tiny$\pm$2.4} \end{tabular}}& 
\multicolumn{1}{c}{\begin{tabular}{c} V: 38.5{\tiny$\pm$0.9} \\ A: 30.3{\tiny$\pm$1.5} \end{tabular}} 
& \multicolumn{2}{c}{\begin{tabular}{c} V: 54.1{\tiny$\pm$3.7} \\ A: 53.7{\tiny$\pm$0.6} \\ T: 72.1{\tiny$\pm$3.3} \end{tabular}}
& \multicolumn{2}{c}{\begin{tabular}{c} V: 64.8{\tiny$\pm$0.2} \\ A: 64.4{\tiny$\pm$0.2} \\ T: 78.9{\tiny$\pm$1.7} \end{tabular}}   
& \multicolumn{1}{c}{\begin{tabular}{c} V: 61.4{\tiny$\pm$0.2} \\ OF: 50.8{\tiny$\pm$0.1} \end{tabular}} \\
\midrule
Ensemble                            &71.7{\tiny$\pm$2.2} & 84.6{\tiny$\pm$1.0} & 70.5{\tiny$\pm$0.2} & 88.4{\tiny$\pm$2.2} & 52.8{\tiny$\pm$0.5} & 70.5{\tiny$\pm$2.1} & 67.2{\tiny$\pm$1.5} & 78.4{\tiny$\pm$0.7} & 77.2{\tiny$\pm$0.6} & 64.6{\tiny$\pm$0.2}\\
Joint Training                      &62.6{\tiny$\pm$5.8} & 74.6{\tiny$\pm$2.2} & 66.7{\tiny$\pm$1.5} & 82.2{\tiny$\pm$0.8} & 47.7{\tiny$\pm$1.5} & 73.0{\tiny$\pm$1.3} & 73.6{\tiny$\pm$1.3} & 80.5{\tiny$\pm$0.2} & 80.8{\tiny$\pm$0.3} & 57.5{\tiny$\pm$0.1}\\
Multi-Loss                          &69.2{\tiny$\pm$1.8} & 82.6{\tiny$\pm$0.9} & 70.1{\tiny$\pm$0.9} & 86.3{\tiny$\pm$1.1} & 51.1{\tiny$\pm$1.8} & 72.1{\tiny$\pm$0.4} & 73.6{\tiny$\pm$2.9} & 80.0{\tiny$\pm$0.7} & 80.2{\tiny$\pm$0.5} & 61.5{\tiny$\pm$0.1}\\
Uni-Pre Frozen                      &72.4{\tiny$\pm$1.8} & 85.0{\tiny$\pm$1.8} & 72.2{\tiny$\pm$0.3} & 87.2{\tiny$\pm$2.4} & 53.0{\tiny$\pm$0.9} & 73.3{\tiny$\pm$1.8} & 72.7{\tiny$\pm$1.6} & 79.9{\tiny$\pm$0.5} & 79.8{\tiny$\pm$0.3} & 64.0{\tiny$\pm$0.2}\\
Uni-Pre Finetuned                   &73.3{\tiny$\pm$1.8} & 82.4{\tiny$\pm$2.0} & 72.5{\tiny$\pm$1.3} & 86.5{\tiny$\pm$0.8} & 53.5{\tiny$\pm$1.3} & 73.1{\tiny$\pm$2.3} & 73.7{\tiny$\pm$0.7} & 80.3{\tiny$\pm$0.4} & 80.3{\tiny$\pm$0.2} & 62.1{\tiny$\pm$0.2}\\
MSLR \cite{MSLR}                    &56.5{\tiny$\pm$2.4} & 77.1{\tiny$\pm$2.4} & 67.3{\tiny$\pm$2.2} & 81.0{\tiny$\pm$1.4} & 50.9{\tiny$\pm$3.9} & $\times$            &$\times$             & $\times$            & $\times$            & $-$\\
MMCosine \cite{xu2023mmcosine}      &59.3{\tiny$\pm$1.5} & 74.0{\tiny$\pm$0.3} & 65.0{\tiny$\pm$1.4} & 83.8{\tiny$\pm$0.8} & 47.3{\tiny$\pm$4.1} & $\times$            &$\times$             & $\times$            & $\times$            & $-$\\
OGM \cite{OGMGE}                    &65.6{\tiny$\pm$3.8} & 82.4{\tiny$\pm$1.0} & 67.3{\tiny$\pm$0.6} & 79.7{\tiny$\pm$1.4} & 51.8{\tiny$\pm$1.9} & 73.9{\tiny$\pm$1.1} & $\otimes$           & 79.7{\tiny$\pm$0.6} & $\otimes$           & 57.8{\tiny$\pm$0.5}\\
AGM \cite{AGM}                      &69.3{\tiny$\pm$1.4} & 78.5{\tiny$\pm$1.6} & 68.4{\tiny$\pm$1.1} & 85.3{\tiny$\pm$0.5} & 51.0{\tiny$\pm$1.6} & 74.0{\tiny$\pm$1.3} & 73.9{\tiny$\pm$1.9} & 79.3{\tiny$\pm$0.4} & 80.2{\tiny$\pm$0.3} & 56.6{\tiny$\pm$0.4}\\
MLB \cite{MLB}                      &71.9{\tiny$\pm$2.2} & 85.2{\tiny$\pm$0.9} & 71.6{\tiny$\pm$0.2} & 86.7{\tiny$\pm$0.3} & 52.2{\tiny$\pm$1.7} & 72.4{\tiny$\pm$1.7} & 74.2{\tiny$\pm$1.7} & 80.1{\tiny$\pm$0.5} & 80.5{\tiny$\pm$0.4} & 61.6{\tiny$\pm$0.2}\\
ReconBoost \cite{ReconBoost}        &69.0{\tiny$\pm$2.4} & 84.8{\tiny$\pm$1.8} & 68.4{\tiny$\pm$1.7} & 86.1{\tiny$\pm$0.7} & 50.2{\tiny$\pm$4.0} & $\times$           &$\times$              & $\times$            & $\times$            & 56.1{\tiny$\pm$0.3}\\
MMPareto \cite{MMPareto}            &69.0{\tiny$\pm$2.5} & 83.8{\tiny$\pm$0.8} & 73.0{\tiny$\pm$1.3} & 87.4{\tiny$\pm$1.3} & 51.4{\tiny$\pm$2.2} & 73.4{\tiny$\pm$1.0} & 73.7{\tiny$\pm$0.6} &79.3{\tiny$\pm$0.6}  & 79.5{\tiny$\pm$0.8} & 59.2{\tiny$\pm$0.5}\\
D\&R \cite{wei2024diagnosing}       &70.6{\tiny$\pm$1.3} & 85.0{\tiny$\pm$0.4} & 72.3{\tiny$\pm$1.5} & \textbf{91.0{\tiny$\pm$0.7}} & 49.3{\tiny$\pm$1.0} & $\times$& $\times$& $\times$& $\times$& 61.7{\tiny$\pm$0.2}\\
\hdashline
$\operatorname{MCR}$  &\textbf{76.1{\tiny$\pm$1.6}} & \textbf{85.7{\tiny$\pm$0.2}} & \textbf{73.4{\tiny$\pm$0.0}} & 88.8{\tiny$\pm$1.0} &  \textbf{55.2{\tiny$\pm$1.8}} & \textbf{75.2{\tiny$\pm$1.7}} & \textbf{76.5{\tiny$\pm$1.4}} & \textbf{80.8{\tiny$\pm$0.4}} & \textbf{81.1{\tiny$\pm$0.4}} &  \textbf{65.0{\tiny$\pm$0.1}}  \\
\bottomrule[1.5pt]
\end{tabular}
}
\captionsetup{position=below} 
\vspace{1mm}
\caption*{
\footnotesize \raggedright
$\times$ method not applicable to regression tasks;\\
$\otimes$ method not applicable to trimodal inputs; \\
$-$ result not reported.}
\vspace{-5mm}
\label{tab:example}
\end{table}


\textbf{Results:} Table~\ref{table:perf_results_mosei_sth} reports accuracy comparisons across baseline methods on our evaluation datasets. We highlight two key observations:

\begin{enumerate}[label=\alph*), leftmargin=0.5cm]
\setlength\itemsep{0em}
\item Most prior methods, including recent multimodal approaches, fail to outperform simpler alternatives such as Ensembles or unimodal encoders (either frozen or finetuned). While some of these methods partially address modality competition, they often fall short of effectively leveraging multimodal data.
\item $\operatorname{MCR}$ is the only method that consistently surpasses all baselines across datasets, model architectures, number of modalities, and task types (classification and regression), with an exception on AVE with Conformer model. This consistent advantage highlights $\operatorname{MCR}$'s ability to balance modality contributions during training under diverse settings, positioning it as a strong and generalizable approach for multimodal learning.
\end{enumerate}

\subsection{Analysis of Multimodal Error}
\label{sec:error_analysis_main}

\textbf{Methodology:} To understand how $\operatorname{MCR}$ improves over existing methods, we perform a post-hoc error analysis by categorizing each sample based on the correctness of the unimodal predictions. Specifically, we consider four groups: (1) both unimodal models are correct, (2) only the first is correct, (3) only the second is correct, and (4) both are incorrect. This breakdown allows us to compare how different multimodal methods behave across these categories and identify whether gains arise from selective reliance on unimodal cues or from synergistic integration.

\begin{figure}[ht!]
    \centering
        \begin{minipage}[t]{\textwidth}
        \centering
        \includegraphics[width=\textwidth]{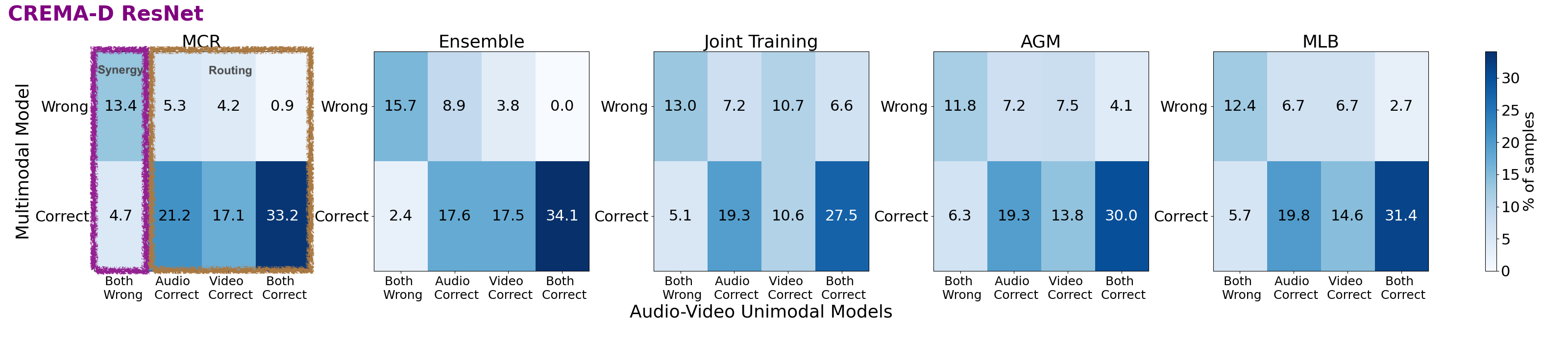}
    \end{minipage}
    \hfill
    \begin{minipage}[t]{\textwidth}
        \centering
        \includegraphics[width=\textwidth]{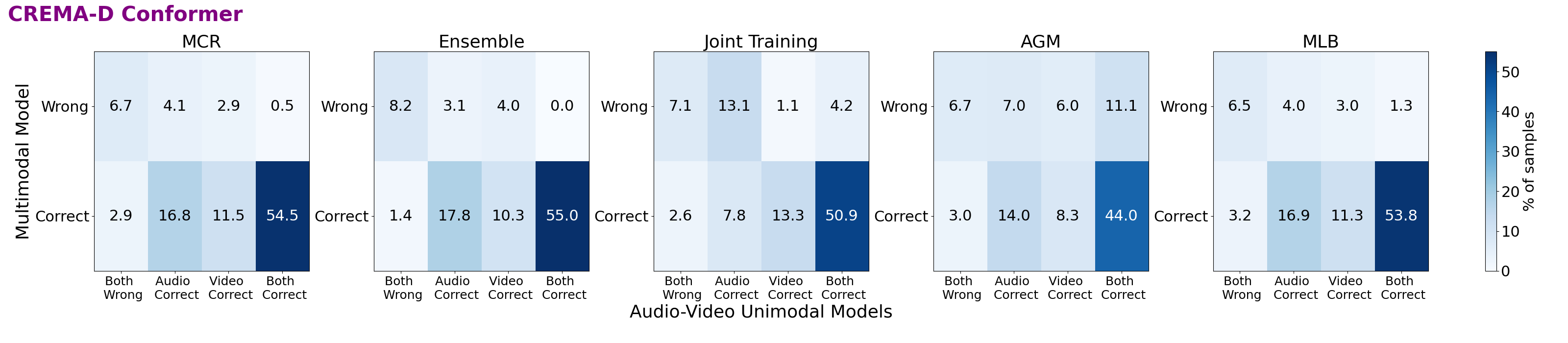}
    \end{minipage}
\vspace{-0.2cm}
\caption{
Error comparison on the CREMA-D dataset across unimodal and multimodal models ($\operatorname{MCR}$, Ensemble, Joint Training, AGM, MLB). Each matrix summarizes model performance based on unimodal prediction correctness. $\operatorname{MCR}$ performs best when at least one unimodal branch is correct (brown box), effectively preserving modality-specific signals. However, AGM and MLB outperform $\operatorname{MCR}$ when both unimodal predictions fail, in the "Both Wrong" (purple box), indicating stronger synergy in those edge cases. Trends across other datasets are shown in Appendix~\ref{app:posthoc_error_analysis}, with MOSI being a notable exception where $\operatorname{MCR}$ also excels in synergy. 
}
\label{fig:posthocerror_main}
\end{figure}

\textbf{Results:} The error analysis of Figure \ref{fig:posthocerror_main} reveals key strengths and limitations of the proposed method. $\operatorname{MCR}$ consistently outperforms other methods in routing information favouring both modalities in the cases that only one of the modalities is correct maintaining competitive performance on both of them and in the case that  all modalities correctly predict the label. This demonstrates $\operatorname{MCR}$'s ability to effectively route information to the appropriate modality, in line with its design choice to model and control training via this modality-independent, task-relevant information through the mutual information terms.

However, $\operatorname{MCR}$ does not exhibit significant gains in capturing synergetic information in the datapoints that all unimodal models fail, underperforming relative to AGM and MLB. This suggests $\operatorname{MCR}$ excels in routing decisions but may be less effective at leveraging synergies across modalities when all individual models falter. This trend holds across other datasets (see Appendix \ref{app:posthoc_error_analysis}), with the exception of MOSI, where $\operatorname{MCR}$ improves synergy. Our initial assumption that concurrent modality training would foster synergy thus did not hold in practice.

\newpage
\section{Discussion}

This paper examines the challenge of modality competition in multimodal learning, where certain modalities dominate the training process, resulting in suboptimal performance. We introduce the Multimodal Competition Regularizer ($\operatorname{MCR}$), a novel approach inspired by information theory, which frames multimodal learning as a game where each modality competes to maximize its contribution to the final output. $\operatorname{MCR}$ efficiently computes lower and upper bounds to optimize both unique and shared task-relevant information for each modality. Our extensive experiments show that $\operatorname{MCR}$ consistently outperforms existing methods and simple baselines on both synthetic and real-world datasets, providing a more balanced and effective multimodal learning framework. $\operatorname{MCR}$ paves the way for fulfilling the long-standing promise of multimodal fusion methods to achieve performance that surpasses the combined results of unimodal training.

We explored different game strategies and observed that directly encouraging competition between modalities in the overall objective function positively impacts performance, as detailed in Appendix \ref{app:ablation_study_strategy}. Future work could investigate more refined strategies to enable individualized and adaptive decisions for each modality to unlock greater performance gains.

Lastly, we conduct a post-hoc error analysis found both in Section \ref{sec:error_analysis_main} and Appendix \ref{app:posthoc_error_analysis}, examining overlaps between the errors of multimodal models and their unimodal counterparts. The results show that $\operatorname{MCR}$ excels at routing decisions to the correct unimodal information but does not promote synergetic behavior accordingly, compared to previous methods. Our initial assumption that the simultaneous progress of unimodal encoders during training would naturally enhance synergy was not supported in practice, highlighting the need for future work to promote this behavior explicitly. Finally, this analysis highlights the potential for performance improvements through enhanced multimodal training, motivating further exploration in this area.




\newpage

\medskip

 {\small
\bibliographystyle{unsrtnat}
\bibliography{citations}  
}




\appendix

\newpage
\section{Supplementary Material}

\subsection{Evidence of limitations of supervised multimodal training}
\label{sec:evidence_of_supervised_limitations}
Identifying the instances where supervised multimodal training collapses is often easier than resolving the issue itself. Nevertheless, it is crucial to understand both these limitations and why simple solutions might not suffice. To explore this, we utilize a ResNet-18 \cite{he2015deep} backbone on the CREMA-D dataset, employing audio and video as the two modalities. These modalities are concatenated just before the final linear layer. We measure multimodal performance at the end of each epoch and, simultaneously, perform linear probing on each modality to evaluate their individual contributions throughout training. 

\begin{figure}[ht]
    \centering
    \includegraphics[width=0.5\linewidth]{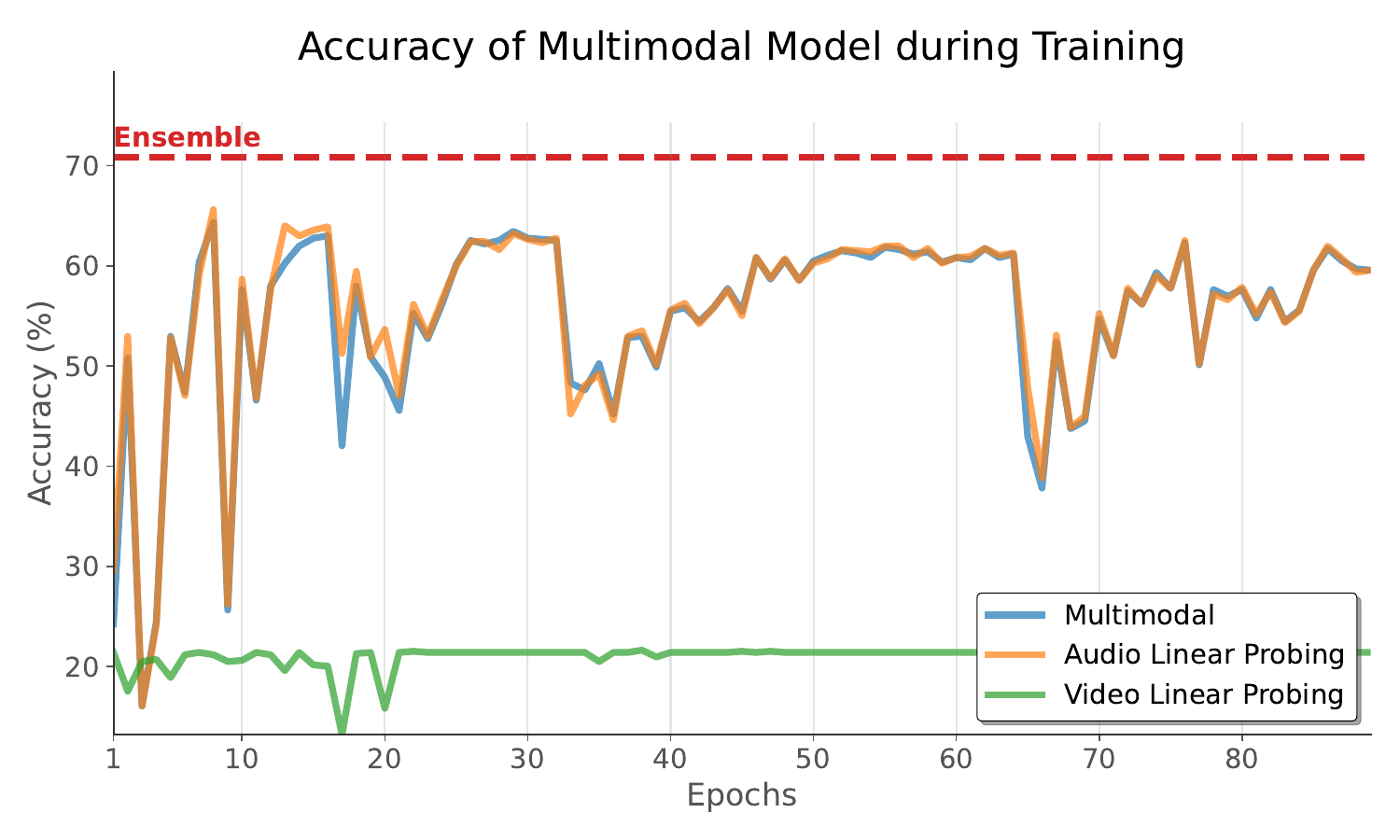}
    \caption{Accuracy of the multimodal model on the CREMA-D dataset across training epochs, showing the performance of the full multimodal model (blue) and individual modality linear probing for audio (orange) and video (green). The dashed red line represents the accuracy of a unimodal ensemble model, highlighting how the model's over-reliance on the audio modality negatively impacts the utilization of the video modality}
    \label{fig:evidence_of_supervised_limitations}
\end{figure}

In Figure \ref{fig:evidence_of_supervised_limitations}, it is clear that the performance of the multimodal model aligns closely with that of the audio modality alone, suggesting that the model heavily relies on audio while neglecting the video modality. This lack of exploration results in the video modality remaining at chance-level accuracy throughout training. As a result, the model fails to leverage any information available in the video modality and performs significantly worse than an ensemble of the unimodally trained models.

\subsection{Multimodal Competition Error}
\label{app:MCE}

Multimodal competition occurs when the network primarily optimizes for one modality, leading to a decline in generalization. One modality dominates the reduction in training error and limits gradient feedback to the other modalities. \citet{huang2022modality} proved that in late fusion models (i.e.\ $\theta_c = \emptyset$), each modality has a probability of causing multimodal competition. They define the phenomenon in terms of the correlation $\Gamma_m= \sum_{cl} \max[\langle X^{cl}_m, W^{cl} \rangle]^+$ of each input modality and the weights, $W$, of the output layer for each corresponding class $cl$. If a modality dominates the competition, it causes the other modality to maintain its initial (i.e.\ before training) correlation levels. Such correlation levels do not have a specific target, unlike other problems where equality is the goal \cite{kariotakis2024}, making the solution less straightforward. We proceed with this analysis on late-fusion models by introducing the generalization error \(\epsilon\) resulting from the effects of multimodal competition.

\begin{definition}
\textbf{Definition of Multimodal Competition Error (MCE):} Let \(\Gamma_1\) and \(\Gamma_2\) represent the correlations of modalities \(Z_1\) and \(Z_2\) with the output. If, after training, \(\Gamma_1 \gg \Gamma_2\), and modality \(X_2\) has predictive power greater than randomness (\(R(X_2) < R(\text{random})\)), while making errors independent from those of \(X_1\) (\(\mathbb{E}[e_1 \cdot e_2] = 0\), where \(e_1\) and \(e_2\) are the errors of \(X_1\) and \(X_2\), respectively), then for a trained model \(f\) and its optimal solution \(f^*\), there exists a multimodal competition error \(\epsilon\), such that:
\begin{equation}
    \epsilon \geq R(f^*) - R_{\text{emp}}(f^*) \text{ and } \epsilon \leq R(f) - R_{\text{emp}}(f)\,
\end{equation}
where \(R(\cdot)= \mathbb{E}[\mathcal{L}_{\text{task}}(Y, \cdot )]\) and \(R_{\text{emp}}(\cdot)\) denote the generalization and empirical model risks, and $\mathcal{L}_{\text{task}}$ represents the corresponding task loss, which could vary depending on the objective, such as cross-entropy for classification or mean-squared error for regression. A higher $\epsilon$ indicates a stronger effect of multimodal competition, implying that the dominance of one modality significantly impacts the model's generalization. These inequalities are empirically observed through improvements in generalization achieved by adjusting the training objectives to address multimodal competition, without altering the model or the data. Numerically estimating $\epsilon$ would require knowledge of the optimal solution \(f^*\), which is typically unavailable. Lastly, transitioning from the late-fusion models of \citet{huang2022modality} we need to extend this definition by replacing correlation with $\operatorname{MI}$ to consider also non-linear statistical dependencies.
\end{definition}

\subsection{Datasets}
\label{app:dataset}
\textbf{CREMA-D \cite{cao2014crema}:} is an emotion recognition dataset with audio and video modalities. It features a diverse group of 91 actors, covering a wide range of ages, ethnicities, and genders. To ensure consistency, each actor is positioned at an equal distance from the camera, expressing six distinct emotions: Happy, Sad, Anger, Fear, Disgust, and Neutral. In alignment with methodologies from prior studies \cite{OGMGE, AGM, PMR}, video frames are sampled at 1 fps, selecting 3 consecutive frames, while audio segments are sampled at 22 kHz, capturing 3 seconds that correspond with the chosen video frames. Audio analysis utilizes a window size of 512 and a step size of 353 samples for Short-Time Fourier Transform (STFT), creating log-Mel spectrograms. For advanced models, our methodology aligns with \citet{Goncalves_2023_2}, incorporating audio signals sampled at 16 kHz and utilizing pre-calculated facial features. Unlike previous approaches \cite{OGMGE, AGM, PMR}, our dataset division follows \citet{Goncalves_2023_2}, excluding actor overlap between training, validation, and test sets. We report standard deviation (std) across folds for consistency.

\textbf{AVE \cite{tian2018audio}:} contains 4143 videos across 28 event categories with a wide range such as frying food or playing guitar, each with temporally labeled audio-visual events of at least 2 seconds. Following \cite{PMR}, video segments where the event occurs are sampled at 1 fps for 4 frames, with audio resampled at 16 kHz using CREMA-D's STFT settings. AVE provides predefined training, validation, and test splits. Our std is derived from three random seeds on the same test set.

\textbf{UCF101 \cite{soomro2012ucf101}:} features real-life action videos from YouTube in 101 action categories, expanding on UCF50. We select samples that include both video and audio modalities, narrowing our focus to 51 action categories. Data preparation mirrors that of the AVE dataset. For model evaluation, we utilize the 3-fold split offered \cite{soomro2012ucf101}, reporting the std across these folds.

\textbf{Something-Something (V2) \citep{goyal2017something}:} presents 220,847 video clips where individuals execute 174 distinct, object-agnostic hand actions, spanning a wide range of simple hand movements without reliance on specific objects. This extensive collection markedly exceeds the data volume of prior datasets. In alignment with \citet{radevski2023multimodal}, we integrate Optical Flow (OF) as the additional modality for capturing dynamic motions. We sample videos at 1 fps, retaining 16 frames per sample. Our data preparation adheres to the pipeline outlined in \citet{radevski2023multimodal}. Results are reported based on three random seeds on the same validation set.

\textbf{CMU-MOSI \citep{zadeh2016mosi} and -MOSEI \cite{zadeh2018multimodal}:} datasets serve as benchmarks for multimodal sentiment and emotion analysis. MOSI comprises 2,199 video clips with video, audio and text modalities. MOSEI expands on this with around 10x more YouTube movie review clips from 1000 different speakers. Both datasets are annotated with sentiment scores (-3 to 3) and following previous works \cite{tsai2019multimodal} we employ metrics such as 7-class accuracy, binary accuracy, F1 score, mean absolute error, and correlation with human annotations for model evaluation. We use for both datasets the aligned versions following \cite{liang2021multibench}. The reported std is derived from three random seeds on the same test set.



\subsection{Models: Backbone Unimodal Encoders}
\label{sec:unimodal_encoders} 

In line with previous research \cite{AGM, OGMGE, PMR, xu2023mmcosine}, our initial experiments adopt ResNet-18 \cite{he2015deep} as the unimodal encoder for handling both video and audio modalities in the CREMA-D, AVE, and UCF datasets. These models are randomly initialized and incorporate adaptive pooling to accommodate diverse input dimensions. For the CMU-MOSEI dataset, we exploit a 5-layer Transformer \cite{vaswani2017attention} similar to previous works \cite{liang2024factorized, liang2021multibench}.

We extend our investigation to include a larger, pre-trained set of unimodal encoders that are optimally suited for each specific modality. This selection process is designed to rigorously assess whether state-of-the-art models exhibit susceptibility to the same phenomena under investigation. We exploit these encoders on the CREMA-D dataset. Following \cite{goncalves2023versatile} on CREMA-D, we deploy the first 12 layers of the Wav2Vec2 \cite{baevski2020wav2vec, wolf-etal-2020-transformers} model with self-supervised pretrained weights for speech recognition, allowing the Wav2Vec2 model to be finetuned. For the video modality, we extract the facing bounding boxes with the multi-task cascaded convolutional neural network (MTCNN) face detection algorithm \cite{zhang2016joint} and afterward the facial features of every available frame exploiting EfficientNet-B2 \cite{tan2019efficientnet} as a frozen feature descriptor. The extracted audio features and pre-calculated facial features are further refined using a 5-layer Conformer \cite{gulati2020conformer}, initialized from scratch. The total model size is 183M parameters. Although the model includes additional components, we refer to it as "Conformer" for simplicity.

For the AVE dataset, we use a similar architecture to CREMA-D, where each branch utilizes an advanced, pretrained model aligned with AVE data, followed by a 5-layer Conformer. For the video branch, we use the ViViT model \cite{arnab2021vivit} pretrained on Kinetics \cite{kay2017kinetics}, and for the audio branch, the HuBERT \cite{hsu2021hubert} model pretrained on Audioset \cite{gemmeke2017audio} resulting in a model with 232M parameters. Ensuring the audio pretraining included non-speech data was important for the pretraining to be beneficial. We also refer to this model as "Conformer," although it incorporates different large pretrained models in this instance.

In the case of the Something-Something dataset, our methodology builds upon the insights presented by \citet{radevski2023multimodal}, which highlight the importance of modality-specific processing in multimodal tasks. For both video and optical flow data, we adopt the Swin Transformer \cite{liu2021swin} as the backbone encoder to each modality. This state-of-the-art architecture excels in capturing hierarchical and spatiotemporal features through its shifted window attention mechanism. 

\subsection{Experimental Details}
\label{sec:experimental_details} 

In this section, we outline the necessary details to reproduce our experiments. Across all datasets, we follow a consistent procedure: we first determine an appropriate learning rate (lr) for the unimodal models by testing several candidates until finding one that works across both modalities. While this step could be avoided by exploiting parameter-specific learning rates, we expected stability implications which we aimed to avoid. For all the experiments of the same dataset/model pair, we use the same hyperparameters, except when fine-tuning pretrained encoders, where we apply a learning rate scaled one magnitude lower. 

All models are optimized using Adam \cite{kingma2014adam} with a cosine learning rate scheduler and a steady warm-up phase, except for the Something-Something dataset, where we use Adaw \cite{loshchilov2017decoupled}. Early stopping is applied for all models, with maximum epochs set to 100 for ResNet and Transformer models, 50 for Conformer models, and 30 epochs in total for Swin Transformers without early stopping. Batch sizes are adjusted based on computational resources, with ResNets and Transformers both using a batch size of 32, Conformers using 8, and Swin-TF using 16. These settings ensure balanced performance and efficient training across all experiments. In several instances, initializing the encoders with pre-trained weights from unimodal training proved beneficial. This was particularly effective for datasets without precomputed features and models without access to larger-scale SSL pretraining. We use different learning rates (\(lr\)) and weight decay (\(wd\)) values across experiments, tailored to each dataset and model. For ResNet models, we use \(lr=1\text{e}{-3}\) and \(wd=1\text{e}{-4}\) for CREMA-D, while both AVE and UCF use \(lr=1\text{e}{-4}\) and \(wd=1\text{e}{-4}\). For Transformer models, including MOSI and MOSEI on two and three modalities, the hyperparameters are consistent \(lr=1\text{e}{-4}\) and \(wd=1\text{e}{-4}\). Similarly, for Conformer models, we set \(lr=5\text{e}{-5}\) and \(wd=5\text{e}{-6}\) for CREMA-D, while AVE uses \(lr=1\text{e}{-4}\) and \(wd=1\text{e}{-4}\). Finally, for Swin-TF models trained on the Something-Something dataset, we configure \(lr=1\text{e}{-4}\) and \(wd=0.02\).

Each of the previous methods includes its own set of hyperparameters, typically just one, with some exceptions such as MSLR or D\&R, which requires additional parameters. For each dataset/model combination, we conduct a brief hyperparameter search, ensuring an equitable number of trials across methods. Due to the extensive list of hyperparameters, we will provide detailed configurations for each experiment in our GitHub repository. The repository link will be included here following the double-blind review process.

Lastly, all of our experiments run on single GPU with different nodes being used for different experiments. For the largest ones we utilized H100 with 80Gb vram to run the experiments of Sth-Sth which required the longest of all up to 48 hours per run. For the rest, we would have from some minutes on the smallest experiment up to 4-5 hours for the datasets CREMA-D, AVE and UCF depending on the GPU and the available RAM.

\subsection{Bounding task-irrelevant information via CEB}
\label{sec:app_ceb}

Our objective is to maximize the information shared between modalities that is irrelevant to the supervised task. Directly estimating the conditional mutual information $I(X_1; X_2 \mid Y)$ is challenging in high-dimensional settings, so we relax the objective by penalizing both the individual and shared irrelevant information, leading to the decomposition:
\begin{align}
I(X_1; X_2 \mid Y) = H(X_1 \mid Y) + H(X_2 \mid Y) - H(X_1, X_2 \mid Y) \\
\Rightarrow - I(X_1; X_2 \mid Y) + H(X_1 \mid Y) + H(X_2 \mid Y) = H(X_1, X_2 \mid Y)
\end{align}

We lower-bound the conditional joint entropy using CEB \cite{fischer2020conditional} as follows,
\begin{align}
H(X_1, X_2 \mid Y) &= -\mathbb{E}_{p(x_1, x_2, y)} \left[ \log p(x_1, x_2 \mid y) \right] \\
&= -\mathbb{E}_{p(x_1, x_2, y)} \left[ \log g(x_1, x_2 \mid y) \right] - \mathrm{KL}\big(p(x_1, x_2 \mid y) \,\|\, g(x_1, x_2 \mid y)\big) \\
&\leq -\mathbb{E}_{p(x_1, x_2, y)} \left[ \log g(x_1, x_2 \mid y) \right]
\end{align}



Assuming a conditional Gaussian model $g(x_1, x_2 \mid y) = \mathcal{N}((x_1, x_2); \mu(y), \sigma^2 I)$, we define $\mu(y)$ as a deterministic function $h: Y; \theta_h \rightarrow Z = (Z_1, Z_2)$ that predicts a target joint representation from $Y$. Then, the conditional entropy term is upper-bounded as:
\begin{equation}
H(X_1, X_2 \mid Y) \leq -\mathbb{E}_{p(x_1, x_2, y)} \left[ \log g(x_1, x_2 \mid y) \right] \propto \mathbb{E}_{p(x_1, x_2, y)} \left\| [f(x_1), f(x_2)] - h(y; \theta_h) \right\|^2
\end{equation}

Thus, we approximate the entropy term with an MSE loss, encouraging $(x_1, x_2)$ to deviate from any deterministic function when it is not informative for the task. Maximizing $H(X_1 \mid Y)$ and $H(X_2 \mid Y)$ incentivizes each modality to retain information that is task-independent, aligning with the goal of minimizing $I(X_1; X_2 \mid Y)$ and suppressing task-irrelevant alignment between modalities.

\subsection{MCR on M modalities}
\label{sec:MIPD_N_Mod}


In this section we provide the analysis of MCR for M number of modalities. In that case, the total mutual information \( I(X_1, \dots, X_M; Y) \) can be decomposed into contributions from individual modalities and their subsets as:
\begin{equation}
I(X_1, .., X_M; Y) = \hspace{-10mm} \sum_{\mathcal{S} \subseteq \{X_1, .., X_M\}, \mathcal{S} \neq \emptyset} \hspace{-10mm} I(\mathcal{S}; Y \mid \{X_1, \dots, X_M\} \setminus \mathcal{S}) + I(X_1, .., X_M) - I(X_1, .., X_M \mid Y),
\end{equation}
where \(\mathcal{S} \subseteq \{X_1, .., X_M\}\) represents a subset of all modalities, excluding the empty set (\(\mathcal{S} \neq \emptyset\)). The term \(\{X_1, .., X_M\} \setminus \mathcal{S}\) denotes the complement of \(\mathcal{S}\), capturing the set of modalities not included in \(\mathcal{S}\). The mutual information \( I(\mathcal{S}; Y \mid \{X_1, .., X_M\} \setminus \mathcal{S}) \) quantifies the information shared between the subset \(\mathcal{S}\) and the target variable \(Y\), conditioned on the remaining modalities. This formulation ensures that all modalities, along with their combinations, are accounted for in the summation. It comprehensively captures interactions at every granularity, from individual modalities (\( |\mathcal{S}| = 1 \)) to the full set of modalities (\( |\mathcal{S}| = M \)). 

While we experimented with reducing the number of terms by considering only the cases where \( |\mathcal{S}| = 1 \), we observed a slight improvement in performance when including all terms for three modalities. However, as the number of modalities increases, it might be beneficial to sub-select and exclude certain terms to mitigate the computational burden and prevent an overflow of terms.


\subsection{Ablation Study - Game Strategies}
\label{app:ablation_study_strategy}

We perform an ablation study to test our hypothesis that framing multimodal competition regularization as a game benefits the model by avoiding destructive loss interactions in each backbone encoder. Table \ref{table:ablation_strategies} compares among the three strategies: Collaborative, Independent, and Greedy. The results show that, across the models allowing backbone encoders to maximize their own $\operatorname{CMI}$ term and concurrently minimizing the others (Greedy strategy) consistently yields the best performance. This result demonstrates that framing multimodal models as competing modalities using game-theoretic principles in the loss terms can be beneficial in balancing these loss terms.
\begin{table}[ht]
\renewcommand{\arraystretch}{1.3}
\centering
\caption{ Ablation Study: Game Strategies – Comparison of model accuracy on CREMA-D dataset for different game strategies: Collaborative, Independent, and Greedy, using both ResNet and Conformer backbones. 
}
\label{table:ablation_strategies}
\begin{small}
\begin{tabular}{lccc}
         & Collaborative  & Independent & Greedy \\
                Dataset/Model Setting &         $\min\limits_{\theta} \mathcal{L}_{\operatorname{MIPD}}$  &    $\min\limits_{\theta_i} \mathcal{L}_{\operatorname{MIPD}_{X_{i}}}$ &       $\min\limits_{\theta_i} \max\limits_{\theta_{\neg i}} \mathcal{L}_{\operatorname{MIPD}_{X_{i}}}$ \\
        
        \toprule
        CREMA-D ResNet + $\operatorname{MCR}$ & 73.4{\tiny$\pm$3.0} & 76.0{\tiny$\pm$2.0} & \textbf{76.2{\tiny$\pm$1.7}} \\
        CREMA-D Conformer + $\operatorname{MCR}$& 82.9{\tiny$\pm$0.7}& 82.6{\tiny$\pm$2.6} & \textbf{85.7{\tiny$\pm$0.2}}\\ 
        AVE ResNet + $\operatorname{MCR}$ & 67.9{\tiny$\pm$2.5} & 72.5{\tiny$\pm$1.0} & \textbf{73.4{\tiny$\pm$0.0}} \\
        AVE Conformer + $\operatorname{MCR}$& \textbf{88.9{\tiny$\pm$1.2}} & 88.6{\tiny$\pm$1.1} & 88.8{\tiny$\pm$1.0} \\
        UCF ResNet + $\operatorname{MCR}$ & 55.1{\tiny$\pm$0.1} &  54.8{\tiny$\pm$1.6} & \textbf{55.2{\tiny$\pm$1.8}} \\ 
        MOSI V-T TF + $\operatorname{MCR}$& {73.7\tiny$\pm$1.3} & {73.6\tiny $\pm$1.7} & \textbf{{75.2\tiny$\pm$1.7}} \\
        MOSI V-A-T TF + $\operatorname{MCR}$& {75.7\tiny$\pm$2.1} & {74.4\tiny$\pm$1.1} & \textbf{{76.5\tiny$\pm$1.4}} \\
        MOSEI V-T TF + $\operatorname{MCR}$& 80.4{\tiny$\pm$0.5} & 80.4{\tiny$\pm$0.5} & \textbf{{80.8\tiny$\pm$0.4}} \\
        MOSEI V-A-T TF + $\operatorname{MCR}$& 80.7{\tiny$\pm$0.2} & 80.8{\tiny$\pm$0.2} & \textbf{{81.1\tiny$\pm$0.4}} \\
        Sth-Sth SwinTF + $\operatorname{MCR}$& 64.9{\tiny$\pm$0.1} & 64.9{\tiny$\pm$0.1} & \textbf{65.0{\tiny$\pm$0.1}} \\
\bottomrule
    \end{tabular} 
\end{small}
\end{table}
\subsection{Ablation Study - Loss Components}
\label{app:ablation_study_loss}

Table \ref{table:ablation_losses} presents the model's performance comparison when different loss components are applied. The models utilize pretrained initialization: ResNet with unimodal pretraining and Conformer with SSL. Two key observations can be made:
\begin{enumerate}[leftmargin=0.5cm]
    \item The concurrent exploitation of both $\mathcal{L}_{\operatorname{MIPD}}$ and $\mathcal{L}_{\operatorname{Con}}$ is yielding consistent improvement. Exploiting them separately leads to smaller improvement for $\mathcal{L}_{\operatorname{Con}}$ and even to a decline for $\mathcal{L}_{\operatorname{MIPD}}$. suggests that alignment in the latent space between the modalities is necessary for the permutations to be effective.

    \item The $\mathcal{L}_{\operatorname{CEB}}$ term, which penalizes task-irrelevant information, improves the Conformer model’s performance, likely due to its pretraining on large, unlabelled datasets that introduce such irrelevant information. In contrast, for the ResNet model, where pretraining already focuses on task-related information, the $\mathcal{L}_{\operatorname{CEB}}$ term does not provide additional benefits.
\end{enumerate}

\begin{table}[h]
\caption{Ablation Study: Regularization Components – Accuracy (\%) of ResNet and Conformer models across datasets with different combinations of $\operatorname{MCR}$ components: $\mathcal{L}_{\operatorname{MIPD}}$, $\mathcal{L}_{\operatorname{Con}}$, and $\mathcal{L}_{\operatorname{CEB}}$.
Results indicate that combining $\mathcal{L}_{\operatorname{MIPD}}$ and $\mathcal{L}_{\operatorname{Con}}$ is crucial for improvement, while $\mathcal{L}_{\operatorname{CEB}}$ does not benefit all models.
}
\centering
\label{table:ablation_losses}
\begin{small}
\centering
\begin{tabular}{lll|lll|ll|l}
         \multicolumn{3}{c}{$\operatorname{MCR}$ Components} &    \multicolumn{3}{|c}{ResNet} & \multicolumn{2}{|c}{Conformer} & \multicolumn{1}{|c}{SwinTF}\\
         $\mathcal{L}_{\operatorname{MIPD}}$ & $\mathcal{L}_{\operatorname{Con}}$ & $\mathcal{L}_{\operatorname{CEB}}$ & CREMA-D & AVE & UCF & CREMA-D & AVE & Sth-Sth \\
        \toprule
        
         & &  & 73.4{\tiny$\pm$2.5} & 71.1{\tiny$\pm$1.4} & 50.0{\tiny$\pm$2.0} & 84.1{\tiny$\pm$0.6} & 87.9{\tiny$\pm$1.1} & 64.7{\tiny$\pm$0.1} \\
         
          \checkmark& & & 74.1{\tiny$\pm$2.9} & 72.1{\tiny$\pm$0.5} & 49.4{\tiny$\pm$1.9} &  83.9{\tiny$\pm$1.8} & 87.8{\tiny$\pm$1.7}& 64.8{\tiny$\pm$0.2} \\
          & \checkmark& & 73.4{\tiny$\pm$2.1} &  72.6{\tiny$\pm$0.6} & 54.8{\tiny$\pm$1.2} & 84.5{\tiny$\pm$0.3} & 88.7{\tiny$\pm$1.4}&  64.7{\tiny$\pm$0.1} \\
         \checkmark& \checkmark&  & \textbf{76.2{\tiny$\pm$1.7}} & \textbf{73.3{\tiny$\pm$0.5}} & \textbf{55.1{\tiny$\pm$0.6}} & 84.5{\tiny$\pm$0.3} & 88.7{\tiny$\pm$1.4}& 64.8{\tiny$\pm$0.1} \\
         
         \checkmark& \checkmark& \checkmark& 75.6{\tiny$\pm$1.9} & 72.1{\tiny$\pm$0.9} & 54.7{\tiny$\pm$1.1} &  \textbf{85.7{\tiny$\pm$0.2}} & \textbf{88.8{\tiny$\pm$1.0}}& \textbf{65.0{\tiny$\pm$0.1}} \\
        \bottomrule
    \end{tabular} 
\end{small}
\end{table}

\subsection{Ablation Study - Perturbation methods}
\label{app:perturbations}

To estimate the importance of a modality, we perturb one modality while keeping the other fixed and observe the change in the model’s output. Several prior methods have proposed ways to do this, but each comes with trade-offs. Additive noise has been used to maximize output variance as a proxy for functional entropy \cite{gat2020removing}, though this increases sensitivity to noise, conflicting with goals such as smoothness and robustness \cite{szegedy2013intriguing, anil2019sorting}. Task-specific augmentations \cite{ji2022increasing, liang2024factorized} rely on handcrafted strategies that may not generalize across domains or modalities. Zero-masking strategies used for approximating Shapley values \cite{AGM} are theoretically grounded but often unreliable in high-dimensional settings \cite{lundberg2017unified} and require multiple forward passes, increasing computational and memory demands.

Based on these previous works we explore three types of perturbation methods to analyze their impact on the performance of $\operatorname{MCR}$: noisy perturbations, zero-masking, and permutations. Each method was applied in different spaces (input space and latent space) or within the batch structure to determine how effectively $\operatorname{MCR}$ can leverage these perturbations to enhance multimodal learning. In Table~\ref{tab:perm_methods}, we summarize the different approaches we examine and in Table~\ref{tab:permutations} we present the results of an ablation study comparing the performance of $\operatorname{MCR}$ under these perturbation techniques across multiple datasets: CREMA-D, AVE, UCF, MOSEI, and MOSI. 

\vspace{-2mm}
\begin{table}[ht!]
\caption{Overview of the approaches examined for the permutation methods.}
\centering
\resizebox{\textwidth}{!}{
\begin{tabular}{ll}
\toprule[1.5pt]
    Noise in the Input Space & Adding noise directly to the input features of each modality,  simulating \\ & realistic data corruption. For its implementation we follow \cite{gat2020removing}. \\
    Shapley Input-Space Perturbations & Following the approach of \cite{AGM}, Shapley zero-induced values \\ & are used to determine the importance of input modalities. \\
    Noise in the Latent Space& Applying noise to the latent representations and encouraging robustness \\ & at the feature extraction level. \\
    Zeros in the Latent Space& Zero-masking latent representations to disrupt one modality. \\
    Within-Batch Permutations in the Latent Space& Permuting data points within the batch to disrupt alignment. \\
\bottomrule[1.5pt]
\end{tabular}}
\label{tab:perm_methods}
\end{table}
\vspace{-2mm}
\begin{table}[ht]
\renewcommand{\arraystretch}{1}
\caption{Ablation study comparing different perturbation methods for $\operatorname{MCR}$ across multiple datasets. The table shows the performance of $\operatorname{MCR}$when combined with various perturbation techniques, including input or latent space noise, Shapley values in the input space, and within-batch permutations.}
\centering
\resizebox{0.85\textwidth}{!}{
\begin{tabular}{cccccc}
\toprule[1.5pt]
Method & CREMA-D & AVE & UCF & MOSEI & MOSI \\ \midrule[1.2pt]
$\operatorname{MCR}$ with Noise Input-Space    & 75.3{\tiny$\pm$2.9} & 72.1{\tiny$\pm$1.1} & 54.6{\tiny$\pm$0.8}          & 80.5{\tiny$\pm$0.4} & 74.7{\tiny$\pm$0.1}\\ 
$\operatorname{MCR}$ with Shapley Input-Space  & 73.6{\tiny$\pm$2.5} & 72.6{\tiny$\pm$0.9} & \textbf{55.5{\tiny$\pm$0.6}} & 79.8{\tiny$\pm$0.5} & 74.3{\tiny$\pm$2.2} \\ 
$\operatorname{MCR}$ with Noise Latent-Space   & 73.6{\tiny$\pm$1.1} & 72.6{\tiny$\pm$0.4} & 54.5{\tiny$\pm$0.7}          & 80.1{\tiny$\pm$0.6} & 72.3{\tiny$\pm$1.7} \\
\conditionalblue{$\operatorname{MCR}$ with Zero Latent-Space}    & 73.6{\tiny$\pm$1.9} & \textbf{73.3{\tiny$\pm$0.5}} & 54.5{\tiny$\pm$0.4} & 79.6{\tiny$\pm$0.4} & 73.6{\tiny$\pm$2.4}\\
$\operatorname{MCR}$ with Permutations Latent-Space & \textbf{76.1{\tiny$\pm$1.1}} & \textbf{73.3{\tiny$\pm$0.5}} & 55.2{\tiny$\pm$1.8} & \textbf{80.8{\tiny$\pm$0.4}} & \textbf{75.2{\tiny$\pm$1.7}} \\
\bottomrule[1.5pt] 
\end{tabular}}
\vspace{-2mm}
\label{tab:permutations}
\end{table}

We observe that Shapley-based input-space perturbations show competitive performance, particularly in datasets like UCF, MOSI, and MOSEI, while noise-based methods (both input and latent spaces) achieve reasonable performance, they consistently underperform other techniques. While input-space perturbations could be a viable option, they significantly increase computational complexity, as they require an additional forward pass through the typically large unimodal encoders for each sample. This limitation, which we analyze in Appendix \ref{app:computational_analysis}, makes them less favorable as a practical solution. Finally, these findings support the choice of permutations as the preferred perturbation method, while suggesting that further exploration of alternative strategies could potentially lead to even greater improvements.

The semantic meaning of this perturbation depends on whether the permuted sample shares the same label as the original. If the labels match, the perturbation is semantically valid and can be seen as an implicit augmentation. In this case, a large output change indicates that the model may be relying on spurious or unstable features within the modality. If the labels do not match, the resulting input is semantically inconsistent and can be interpreted as out-of-distribution. If the model’s output is insensitive to such a perturbation, this suggests that the modality is being ignored. Conversely, a sensitive reaction may indicate an overreliance on features not robust to semantic shifts. Therefore each category of permuted samples contributes differently to the final output. In practice, we apply both semantically consistent and inconsistent permutations during training. This choice introduces minimal computational overhead and appears to slightly improve convergence stability. 

\conditionalblue{
We note that the gradients $\nabla_{\theta_1}\mathcal{L}_{\operatorname{MIPD}_2}$ and $\nabla_{\theta_2}\mathcal{L}_{\operatorname{MIPD}_1}$ can negatively impact model robustness depending on the type of perturbation applied. When perturbations yield out-of-distribution unimodal inputs, such as zero-masking or additive noise, the resulting gradients may encourage the model to learn spurious patterns. In contrast, our main experiments use within-batch permutations, which preserve the in-distribution structure of each modality. Under permutation as the perturbation method, the $\mathcal{L}_{\operatorname{MIPD}}$ formulation remains symmetric regardless of which modality is perturbed, and the gradients contribute constructively to learning in all branches.}

\subsection{Analysis of Multimodal Error}
\label{app:posthoc_error_analysis}

\begin{figure}[t!]
    \centering

    \begin{minipage}[t]{0.99\textwidth}
     \includegraphics[width=\textwidth]{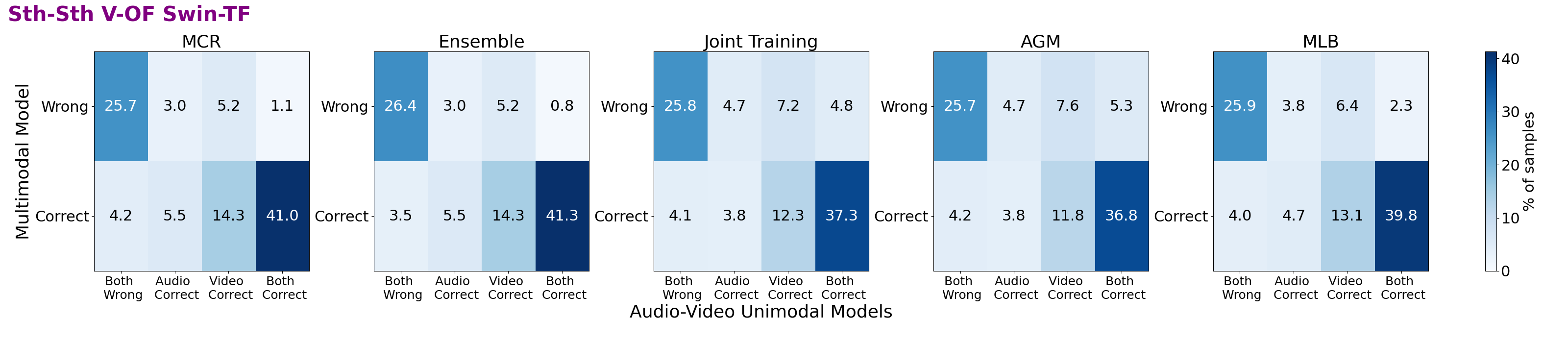}
    \end{minipage}
\vspace{-2mm}
\caption{Error comparison matrices across pairs of datasets and models, comparing unimodal predictions with multimodal models trained using various methods, including Ensemble, Joint Training, AGM, MLB, and $\operatorname{MCR}$ for the datasets AVE, UCF, MOSI, MOSEI and Something-Something (Sth-Sth). Each column of the confusion matrix represents cases where both unimodal predictions are incorrect, where only one is correct, and where both are correct. The results highlight that $\operatorname{MCR}$ consistently performs well in cases where at least one unimodal prediction is correct. Additionally, MLB and AGM in many instances outperform $\operatorname{MCR}$ in discovering synergetic information, which refers to the "Both/All Wrong" column, highlighting a current limitation of $\operatorname{MCR}$.}
\label{fig:conf_matrix}
\end{figure}

We extend the error analysis from Section \ref{sec:error_analysis_main} by comparing unimodal and multimodal predictions in Figure \ref{fig:conf_matrix}. The results echo the pattern seen in CREMA-D (Figure \ref{fig:posthocerror_main}): $\operatorname{MCR}$ excels when at least one unimodal model predicts correctly, but still trails MLB and AGM when all unimodal models fail. An exception is the MOSI dataset, where $\operatorname{MCR}$ performs well in synergy, even with three modalities. 
    
\begin{figure}[b!]
    \centering
    \begin{minipage}[t]{0.99\textwidth}
     \includegraphics[width=\textwidth]{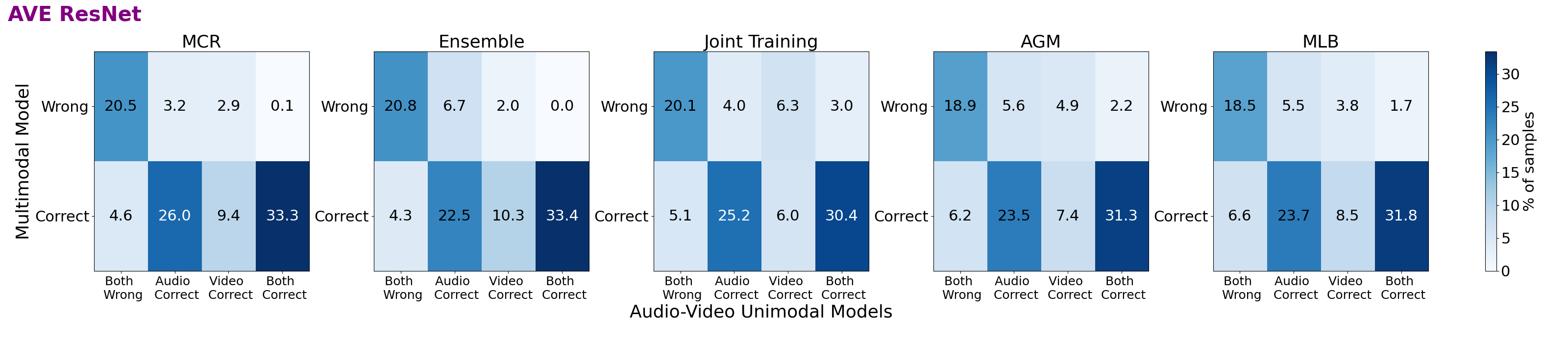}
    \end{minipage}
    \begin{minipage}[t]{0.99\textwidth}
     \includegraphics[width=\textwidth]{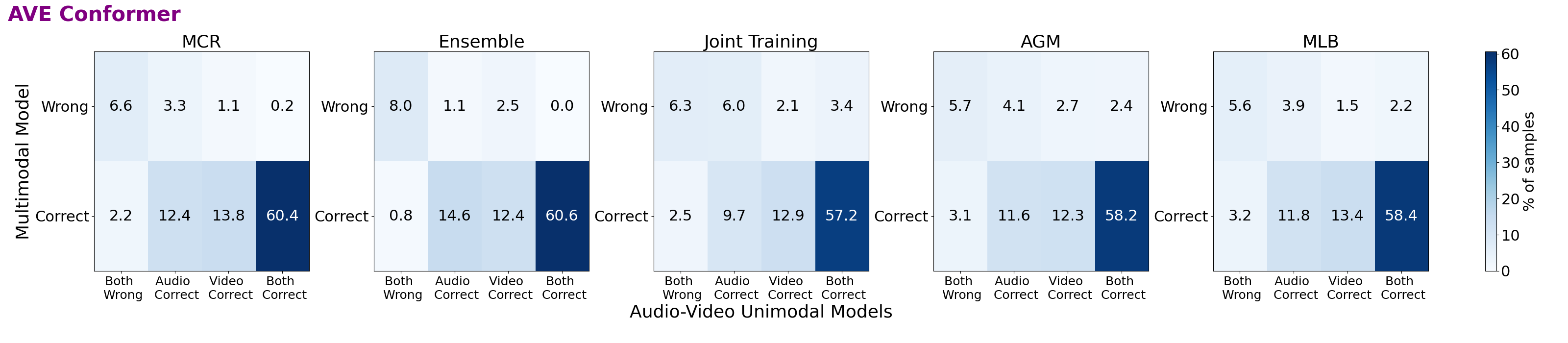}
    \end{minipage}
    \begin{minipage}[t]{0.99\textwidth}
     \includegraphics[width=\textwidth]{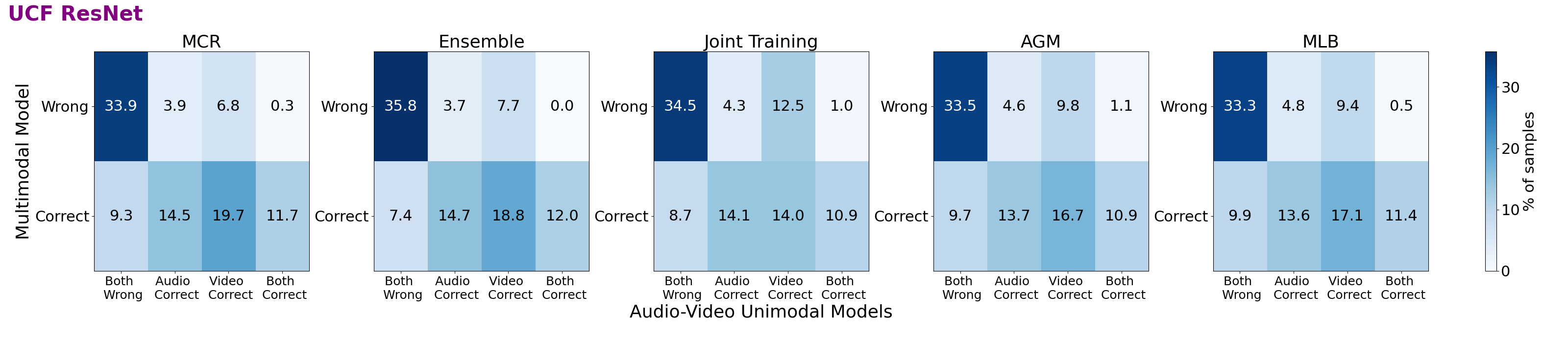}
    \end{minipage}
    \begin{minipage}[t]{0.99\textwidth}
     \includegraphics[width=\textwidth]{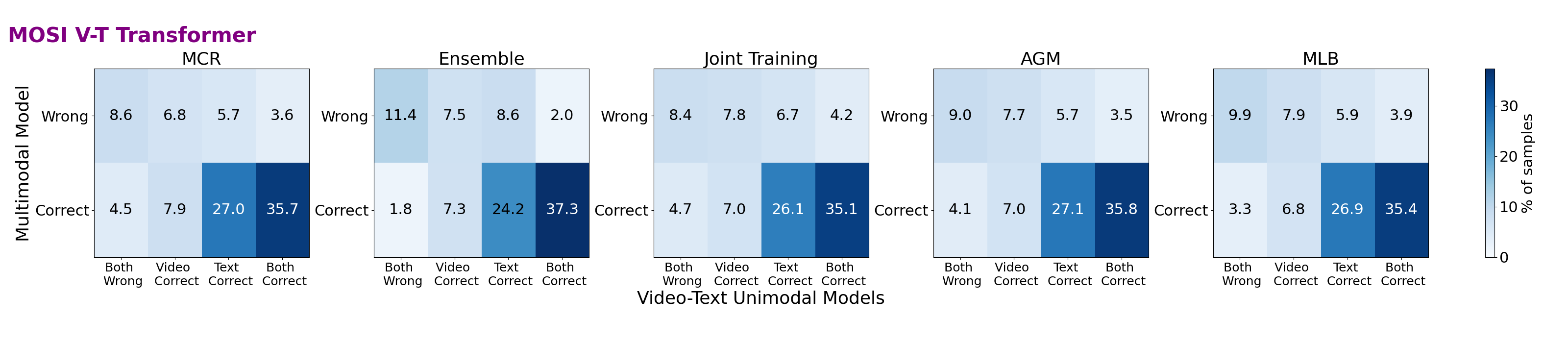}
    \end{minipage}
    \begin{minipage}[t]{0.99\textwidth}
     \includegraphics[width=\textwidth]{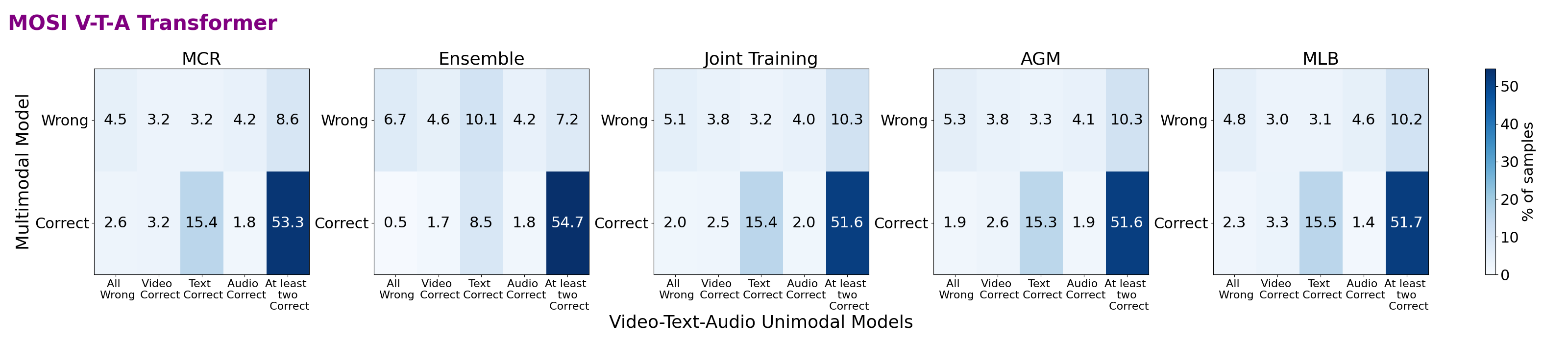}
    \end{minipage}
        \begin{minipage}[t]{0.99\textwidth}
     \includegraphics[width=\textwidth]{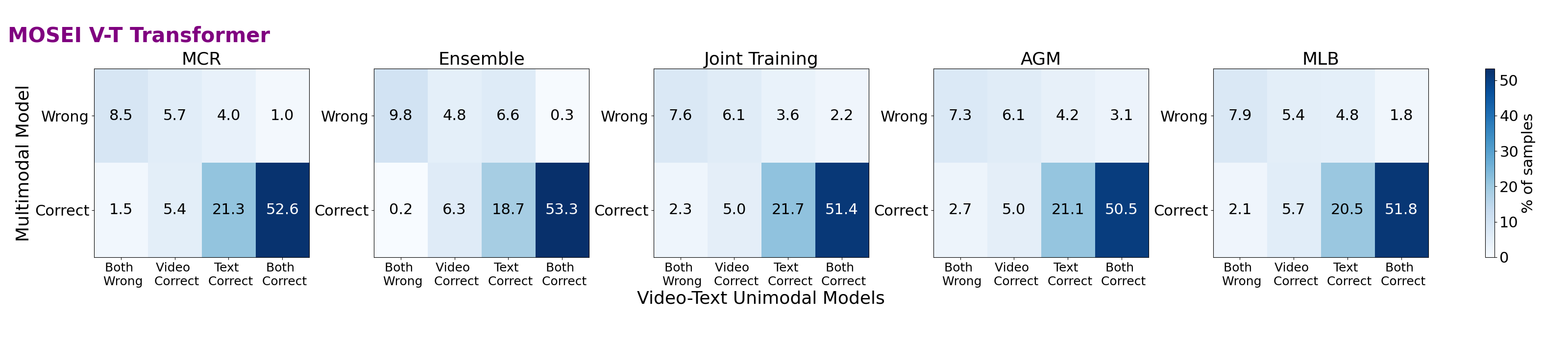}
    \end{minipage}
    \begin{minipage}[t]{0.99\textwidth}
     \includegraphics[width=\textwidth]{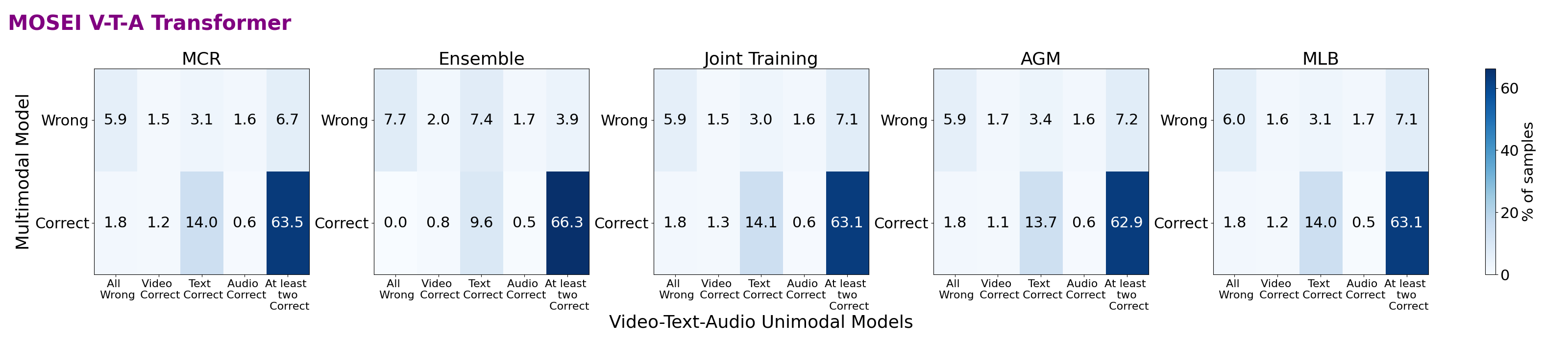}
    \end{minipage}
\end{figure}

\subsection{Statistical Importance}
\label{app:statistical_importance}

We assess the statistical significance of the performance differences between our method and each baseline using the Wilcoxon Signed-Rank Test, applied to per-dataset average results. To control for multiple comparisons, we apply the Holm correction to the resulting $p$-values. We consider results significant if the adjusted $p$-value is below $\alpha = 0.05$. Full results are shown in Table~\ref{tab:statistical_tests}.

\begin{table}[h]
\centering
\caption{Wilcoxon Signed-Rank Test results (two-sided) comparing our method to each baseline across datasets. Statistically significant comparisons after Holm correction ($\alpha = 0.05$) are bolded.}
\label{tab:statistical_tests}
\begin{tabular}{lcc}
\toprule
\textbf{Comparison} & \textbf{Raw $p$-value} & \textbf{Holm-adjusted $p$-value} \\
\midrule
$\operatorname{MCR}$ vs Ensemble & \textbf{0.00195} & \textbf{0.02148}\\

$\operatorname{MCR}$ vs Joint Training & \textbf{0.00195} & \textbf{0.01074}\\

$\operatorname{MCR}$ vs Multi-Loss & \textbf{0.00195} & \textbf{0.00716}\\

$\operatorname{MCR}$ vs Uni-Pre Frozen & \textbf{0.00195} & \textbf{0.00537}\\

$\operatorname{MCR}$ vs Uni-Pre Finetuned & \textbf{0.00195} & \textbf{0.00430}\\

$\operatorname{MCR}$ vs OGM & \textbf{0.00781} & \textbf{0.00955}\\

$\operatorname{MCR}$ vs AGM & \textbf{0.00195} & \textbf{0.00358}\\

$\operatorname{MCR}$ vs MLB & \textbf{0.00195} & \textbf{0.00307}\\

$\operatorname{MCR}$ vs ReconBoost & \textbf{0.03125} & \textbf{0.03438}\\

$\operatorname{MCR}$ vs MMPareto & \textbf{0.00195} & \textbf{0.00269}\\

$\operatorname{MCR}$ vs D\&R & 0.15625 & 0.15625\\
\midrule
Greedy vs Collaborative & \textbf{0.00977} & \textbf{0.00977}\\
Greedy vs Independent & \textbf{0.00195} & \textbf{0.00391}\\

\midrule
$L_{MIPD}$ + $L_{Con}$ + $L_{CEB}$ vs No Reg & \textbf{0.03125} & 0.12500\\
$L_{MIPD}$ + $L_{Con}$ + $L_{CEB}$ vs $L_{MIPD}$ & \textbf{0.04311} & 0.08623\\
$L_{MIPD}$ + $L_{Con}$ + $L_{CEB}$ vs $L_{Con}$ & 0.06789 & 0.09052\\
$L_{MIPD}$ + $L_{Con}$ + $L_{CEB}$ vs $L_{MIPD}$ + $L_{Con}$ & 0.84375 & 0.84375\\
\bottomrule
\end{tabular}
\end{table}

\section*{Dynamics of MIPD Components}
\label{sec:loss_dynamics}

\begin{figure}[h!]
    \centering
    \includegraphics[width=0.9\textwidth]{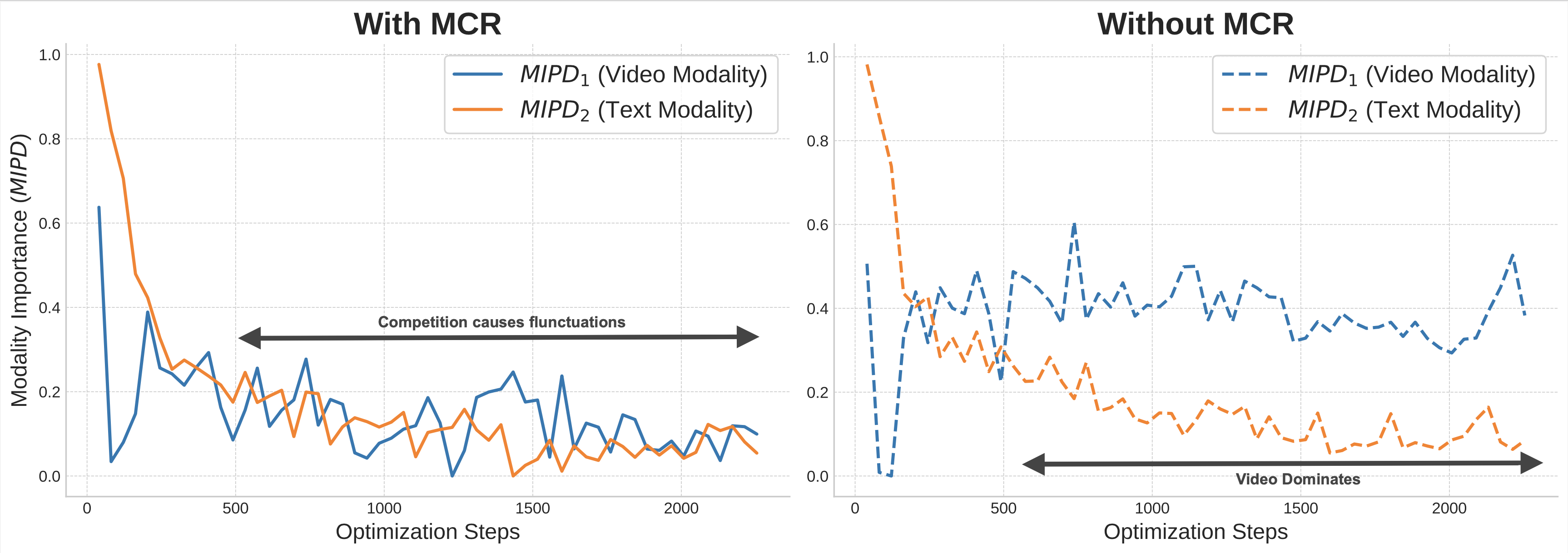}
    \caption{The plots show the two $\operatorname{MIPD}$ losses ($\operatorname{MIPD}_1$ and $\operatorname{MIPD}_2$) for each modality, with (left) and without (right) regularization. With regularization, the losses exhibit an alternating pattern, demonstrating $\operatorname{MCR}$’s ability to dynamically explore multiple modality contribution combinations and prevent collapse into a single modality. Without regularization, the text modality initially contributes more strongly, but the video modality eventually outperforms and dominates.}
    \label{fig:loss_dynamics}
\end{figure}

To illustrate the learning dynamics of the $\operatorname{MCR}$ regularizer, we can analyze the evolution of its core loss components during training. Figure~\ref{fig:loss_dynamics} plots the two $\operatorname{MIPD}$ terms, corresponding to the video and text modalities, from a training run on the MOSI V-T dataset.

These $\operatorname{MIPD}$ terms, which serve as proxies for the unique contribution of each modality, exhibit a dynamic, alternating behavior. The fluctuations reflect shifts in which modality is more influential on the fused output at different stages of training. This plot visualizes $\operatorname{MCR}$'s mechanism for actively balancing modality importance, preventing one from consistently dominating the other. It is important to note that while this alternating pattern is the desired behavior, its specific form and prominence can vary across different training runs and datasets. Therefore, this figure is presented as a clear, illustrative example of the dynamic interplay $\operatorname{MCR}$ encourages. This prevents static dominance by a single modality, which is key to mitigating modality competition.

\newpage
\subsection{Reproduction Challenges for MLA}

We attempted to reproduce the results reported for the MLA method~\cite{zhang2024multimodal} using the authors’ officially released code. Despite carefully replicating the training procedure, we were unable to reach close to the reported performance. Specifically, our reproduced accuracy results on our five different settings are as follows: CREMA-D (ResNet) 61.7{\tiny$\pm$4.1}, CREMA-D (Conformer) 75.1{\tiny$\pm$3.5}, AVE (ResNet) 44.5{\tiny$\pm$1.9}, AVE (Conformer) 72.4{\tiny$\pm$3.5}, and UCF (ResNet) 47.9{\tiny$\pm$1.4}. These results are substantially lower than those reported in Table 1 of the original paper. We suspect the discrepancy may be due to missing implementation details. We include these findings in the interest of transparency and to support ongoing efforts toward reproducibility in multimodal learning.

\subsection{Computational Speed and Memory Analysis of Perturbations}
\label{app:computational_analysis}
The computational load imposed by any sample that requires an additional pass can be divided into the encoders $f_{1}, f_{2}$ and the fusion network $f_c$. The encoders have a computational cost of $cost_{enc}$ per sample, and the fusion network has a cost of $cost_c$ per sample. Thus, the total computational complexity is $\mathcal{O}((M+1)*N*(cost_{enc}+cost_{c}))$, where $M$ is the times we draw noisy samples to which we add the non-perturbed batch and $N$ is the batch size. If we now use permutation samples that can be directly drawn from the latent space, the additional computational complexity is reduced to $\mathcal{O}(N*cost_{enc}+(M+1)*N*cost_{c})$. Compared to the necessity of a supervised forward pass this translates to an addition of $\mathcal{O}(M*N*cost_{c})$ computational burden. In most state-of-the-art models, each modality encoder is significantly larger than the fusion network, resulting in $cost_{enc} >> cost_{c}$. In such networks, permutations can have almost negligible additional computations. The memory footprint follows a similar pattern.

\section{Proof of Supervised Contrastive Loss as Lower Bound}
\label{sec:app_Con_proof}
We consider the supervised contrastive loss with $\psi$ being the critic function and we rewrite it as follows:
\begin{equation}
\label{eq:app_contrastive}
\mathcal{L}_{\text{Con}}(X_1, X_2) = \sum_{i \in \mathbb{I}} \frac{-1}{\left| \mathcal{P}_i \right|} \sum_{k \in \mathcal{P}_i}  \left[ \log \frac{\psi(x_{1_i}, x_{2_k})}{\sum_{j \in \mathbb{I}}  \psi(x_{1_i}, x_{2_j})} \right]
\end{equation}
where $\mathcal{P}_i = \{p \in \mathbb{I} \mid y_p = y_i\}$. In supervised contrastive learning, the presence of multiple positive samples turns this into a multi-label problem, unlike traditional noise contrastive estimation (NCE) methods \cite{oord2018representation}, which typically assume only one positive sample. By taking the version of supervised contrastive learning with the expectation over the positives outside of the log, we can interpret each classification as an average of classifiers, with each classifier focusing on identifying one of the positive samples. 

For each positive sample $p \sim \mathcal{P}_i$, we aim to derive the optimal probability of correctly identifying that point, denoted $d = p$. This is done by sampling the point from the conditional distribution $p(x_{2_p} \mid x_{1_i}, y_i)$ while sampling the remaining points from the proposal distribution $p(x_{2_l})$. This approach mirrors the technique used in InfoNCE \cite{oord2018representation} and leads to the following derivation:
\begin{align}
    p(d=p |X_2, x_{1_i}, y_i) &= \frac{p(x_{2_p} \mid x_{1_i}, y_i) \underset{l \in \mathcal{I}, l \neq p
    }{\Pi} p(x_{2_l}) }{ \sum_{j \in \mathcal{I}}   p(x_{2_j} \mid x_{1_i}, y_i) \underset{l \in \mathcal{I}, l \neq j
    }{\Pi} p(x_{2_l}) }\\
    \label{eq:opt_prob_sup_con}
    &=  \frac{\frac{p(x_{2_p} \mid x_{1_i}, y_i)}{p(x_{2_p})}}
    { \sum_{j \in \mathcal{I}}  \frac{p(x_{2_j} \mid x_{1_i}, y_i)}{p(x_{2_j})} }
\end{align}
The optimal value for the critic function $\psi$ in Equation \ref{eq:opt_prob_sup_con} is proportional to $\psi \propto \frac{p(x_{2_p} \mid x_{1_i}, y_i)}{p(x_{2_p})}$. The MI between the variables can be estimated as follows: 
\begin{align}
    L_{\text{Con}}^{Opt} &= - \underset{i \sim \mathcal{I}}{\mathbb{E}} \left[ \underset{p \sim \mathcal{P}_i}{\mathbb{E}} \log \left[ 
    \frac{ \frac{p(x_{2_p} \mid x_{1_i}, y_i)}{p(x_{2_p})}}{ \frac{p(x_{2_p} \mid x_{1_i}, y_i)}{p(x_{2_p})} + \underset{j \in \mathcal{I}, j \neq i}{\sum}  \frac{p(x_{2_j} \mid x_{1_i}, y_i)}{p(x_{2_j})}}
    \right] \right] \\ 
    &= \underset{i \sim \mathcal{I}}{\mathbb{E}} \left[ \underset{p \sim \mathcal{P}_i}{\mathbb{E}} \log \left[ 1 + 
    \frac{p(x_{2_p})}{p(x_{2_p} \mid x_{1_i}, y_i)} \underset{j \in \mathcal{I}, j \neq i}{\sum} \frac{p(x_{2_j} \mid x_{1_i}, y_i)}{p(x_{2_j})}
    \right] \right]\\
    &= \underset{i \sim \mathcal{I}}{\mathbb{E}} \left[ \underset{p \sim \mathcal{P}_i}{\mathbb{E}} \log \left[ 1 + \frac{p(x_{2_p})}{p(x_{2_p} \mid x_{1_i}, y_i)} (N-1)  \underset{j \in \mathcal{I}}{\mathbb{E}} \frac{p(x_{2_j} \mid x_{1_i}, y_i)}{p(x_{2_j})}
    \right] \right]\\
    &= \underset{i \sim \mathcal{I}}{\mathbb{E}} \left[ \underset{p \sim \mathcal{P}_i}{\mathbb{E}} \log \left[ 1 + \frac{p(x_{2_p})}{p(x_{2_p} \mid x_{1_i}, y_i)} (N-1)
    \right] \right]\\
    &\geq \underset{i \sim \mathcal{I}}{\mathbb{E}} \left[ \underset{p \sim \mathcal{P}_i}{\mathbb{E}} \log \left[ \frac{p(x_{2_p})}{p(x_{2_p} \mid x_{1_i}, y_i)} N
    \right] \right]\\
    &= \log N - I(X_{2};X_{1}, Y), \text{ using MI properties \citep[Chapter~6.3.4]{murphy2022probabilistic}}\\
    &= \log N - I ( X_{2} ; Y |X_{1}) - I(X_{2} ; X_{1})
\end{align}

Therefore, by taking both sides of the contrastive loss to predict $X_2$ from $X_1$ and $X_1$ from $X_2$ we derive to  $I ( X_{2} ; Y |X_{1}) + I (X_{1} ; Y |X_{2}) + 2 \cdot I(X_{2} ; X_{1}) \geq \log N - L_{\text{Con}}^{Opt}$. This trivially also holds for other $\psi$ that obtain a worse(higher) $L_{\text{Con}}$. Simarly to InfoNCE, the bound becomes more accurate as N increases, while due to the the term $\frac{1}{\left| \mathcal{P}_i \right|}$ it is not affected by the number of positive pairs.





\end{document}